\pdfoutput=1
\documentclass[11pt,a4paper]{article}

\usepackage{framed,multirow}
\usepackage{amssymb}
\usepackage[pdftex]{graphicx}
\usepackage{longtable}
\usepackage{amsmath}
\usepackage{latexsym}
\usepackage{booktabs}
\usepackage{pifont}
\usepackage{float}
\usepackage{wrapfig}
\usepackage{subcaption}
\usepackage[numbers]{natbib}
\usepackage{url}
\usepackage{xcolor}
\definecolor{newcolor}{rgb}{.8,.349,.1}
\usepackage{hyperref}
\usepackage[utf8]{inputenc}
\usepackage[a4paper, margin=0.75in]{geometry}

\newcommand{\cmark}{\ding{51}}%
\newcommand{\xmark}{\ding{55}}%

\hypersetup{
    colorlinks=true,
    linkcolor=blue,
    anchorcolor=blue,
    citecolor=blue
}

\begin{document}

\title{CFDBench: A Large-Scale Benchmark for Machine Learning Methods in Fluid Dynamics}
\date{}
\author{
    \textbf{Yining Luo$^{1*}$} \hspace{1cm}
    \textbf{Yingfa Chen$^{2*}$} \hspace{1cm}
    \textbf{Zhen Zhang$^{1\dagger}$} \\
    $^1$Key Laboratory of Advanced Nuclear Reactor Engineering and Safety of\\ 
    Ministry of Education, Institute of Nuclear and New Energy Technology, \\
    Collaborative Innovation Center of Advanced Nuclear Energy Technology\\
    $^2$Department of Computer Science and Technology \\
    Tsinghua University, Beijing, P. R. China\\
    \texttt{\{luoyn21,yf-chen22\}@mails.tsinghua.edu.cn} \\
}



{\let\newpage\relax\maketitle}


\begin{abstract}
In recent years, applying deep learning to solve physics problems has attracted much attention. Data-driven deep learning methods produce fast numerical operators that can learn approximate solutions to the whole system of partial differential equations (i.e., surrogate modeling). 
Although these neural networks may have lower accuracy than traditional numerical methods, they, once trained, are orders of magnitude faster at inference. Hence, one crucial feature is that these operators can generalize to unseen PDE parameters without expensive re-training.
In this paper, we construct CFDBench, a benchmark tailored for evaluating the generalization ability of neural operators after training in computational fluid dynamics (CFD) problems. 
It features four classic CFD problems: lid-driven cavity flow, laminar boundary layer flow in circular tubes, dam flows through the steps, and periodic Karman vortex street. 
The data contains a total of 302K frames of velocity and pressure fields, involving 739 cases with different operating condition parameters, generated with numerical methods.
We evaluate the effectiveness of popular neural operators including feed-forward networks, DeepONet, FNO, U-Net, etc. on CFDBnech by predicting flows with non-periodic boundary conditions, fluid properties, and flow domain shapes that are not seen during training. 
Appropriate modifications were made to apply popular deep neural networks to CFDBench and enable the accommodation of more changing inputs. 
Empirical results on CFDBench show many baseline models have errors as high as 300\% in some problems, and severe error accumulation when performing autoregressive inference.
CFDBench facilitates a more comprehensive comparison between different neural operators for CFD compared to existing benchmarks.\footnote{The code and datasets can be found at: \url{https://www.github.com/luo-yining/CFDBench}}

\end{abstract}


\section{Introduction}

Recent advances in deep learning have enabled neural networks to approximate highly complex and abstract mappings from large-scale data \citep{deep-learning}. This has led to the emergence of \textit{surrogate modeling} as a technique to learn fast and approximate solvers for partial differential equations (PDEs) using neural networks, and this has shown some promising results in various domains \citep{neural-operator,fno,deeponet,pangu-weather}.

One application of PDE solvers is computational fluid dynamics (CFD), which is a well-studied and important field with many practical applications. Therefore, the last few years saw many new attempts at developing better CFD methods with the help of deep neural networks \citep{ml-accelerate-cfd}. These neural models are trained on large-scale data by approximating input-output pairs through gradient descent. There are multiple reasons for adopting deep learning methods over traditional numerical methods. One advantage is mesh-independence. Numerical methods operate on meshes, and the mesh construction process is time-consuming and requires much expert knowledge to ensure convergence and good accuracy. In contrast, some numerical operators are mesh-independent, allowing input and output at arbitrary locations. Additionally, although deep neural networks need to undergo a long training process, with the help of modern hardware, they can be several orders of magnitude faster than numerical methods at prediction \citep{pinn,fno}. Finally, certain neural models have been able to surpass traditional numerical methods in accuracy in some problems in fluid dynamics \citep{metnet,pangu-weather}.

In this paper, we concern ourselves with the ability to generalize to unseen PDE parameters (e.g., different BCs, physical properties, domain geometry, etc.) without re-training. Namely, this work focuses on neural operators that can accept PDE parameters as a part of the inputs along with the input function, and make predictions conditioned on both the input function and the PDE parameters. 
Such ability is important for practical applications of neural operators because the training process is typically costly in terms of time and computational power (often requiring days or even weeks). However, most existing works only evaluate the generalization to unseen initial conditions\footnote{Note that we regard the cases where a neural network learns the mapping from a certain frame of the flow at time step $t$ to a later time step $t + \Delta t$ (i.e., $u(t) \rightarrow u(t+\Delta t)$) as varying initial conditions because $u$ at any time step can be seen as the IC for subsequent steps.} (ICs) \citep{deeponet,fair-comparison-neural-operators,neural-operator}. This lack of a standardized evaluation benchmark on parameter generalization has largely limited our understanding of how different design choices of neural operators affect their effectiveness at generalizing along dimensions other than the ICs.

To address this lack of evaluation, we construct CFDBench, a large-scale dataset for better evaluating the generalization ability of data-driven neural networks in CFD. It includes four classic CFD problems: the flow in a lid-driven cavity, the flow in a circle tube, the flow over a dam, and the flow around a cylinder problem. In contrast to existing work, we condition the neural networks on different BCs, fluid physical properties, and fluid domain geometry, and evaluate their generalization effectiveness to unseen conditions. CFDBench is generated over the course of several days using industry-level proprietary software, which utilizes state-of-the-art numerical methods. The data is interpolated into grid-like features for array-based neural models. We emphasize that numerical methods are likely an upper bound for the performance of the deep learning methods.

Our main contributions are as follows. 

\begin{enumerate}
    \item We construct and release the first benchmark tailored for CFD data-driven deep learning, covering four classic CFD problems with different BCs, fluid properties, and domain geometry.
    \item Some neural networks cannot be directly applied to CFDBench, and we demonstrate how to modify them to be effectively applied to the problems in CFDBench. 
    \item We evaluate some popular neural networks on CFDBench, and show that it is more challenging than many virtual problems used in previous works, revealing some problems that need to be solved before these operators can replace traditional solvers.
\end{enumerate}

\section{Related works}

\subsection{Numerical Methods for Solving PDEs}

Numerical methods have been widely used to solve CFD problems. The basic idea is to divide the original continuous solution area into many finite interconnected regions, enabling using a point (called node) to approximately represent an entire small region. Then, using different discrete methods, the governing equations (which are typically PDEs) can be reduced to algebraic equations called discrete equations. Solving these discrete equations gives us the values of the nodes. Some common discrete methods include finite difference methods, finite volume methods, finite element methods, spectral methods, and lattice Boltzmann methods (LBMs).

The main idea of the finite difference method (FDM) \cite{finite-difference-method} is to approximates derivatives in partial differential equations (PDEs) at grid points, often using Taylor series expansions, converting PDEs into a system of algebraic equations. Although its theory is simple, it struggles to handle complex geometries and irregular boundaries. The finite volume method (FVM) \cite{finite-volume-method} divides the calculation area into non-repeating control volumes and then integrates fluxes across control volumes (typically grid cells), ensuring strict adherence to conservation laws. The finite element method (FEM)\cite{finite-element-method} is based on the classical variational method (Ritz method \cite{ritz-method} or Galerkin method \cite{galerkin-method}), which divides the domain into finite elements and approximates the solution using basis functions within each element, transforming PDEs into algebraic equations. The advantages of the FVM and FEM are good conservation and good adaptability to complex grids. At the same time, the disadvantage is high computing consumption and that its convergence is highly dependent on the quality of the mesh. The spectral method \cite{spectral-method} uses the characteristics of the Fourier series to transform the nonlinear problem into a linear problem. Its advantages include high accuracy and great applicability to problems with periodic BCs, but it has considerable limitations such as divergence on discontinuous functions. Lattice Boltzmann methods (LBMs) \cite{lattice-boltzmann} is a more recent method based on the thin (mesoscopic) scale model and Boltzmann gas molecular motion theory, with the advantage of fast solution speed, however, it often results in lower accuracy as a compromise. Concerning our work, most numerical methods with relatively high accuracy have very large computational costs. Despite many efforts to reduce such computational costs has been made, the costs are still very great in many industrial and commercial applications. Therefore, deep neural models, which can better make use of modern hardware, emerge as a viable workaround for reducing this computational cost.

\subsection{Neural Networks}

In the last decade, neural networks have demonstrated impressive capabilities in various computer vision and natural language processing tasks \cite{deep-learning,prelu,resnet,gpt3,bert}. A neural network consists of a large number of neurons. It can approximate any arbitrary mapping by automatically minimizing a loss function that is differentiable with respect to the model parameters. By iterating through a large set of input-output pairs, the model parameters are updated by gradient descent. Some common types of neural networks include feed-forward neural networks (FFNs), recurrent neural networks (RNNs)\cite{lstm}, generative adversarial networks (GANs)\cite{gan}, convolutional neural networks (CNNs)\cite{neocognitron} etc.

Regarding CFD problems, we generally want to model a flow field, which can be seen as a kind of condition generation task. This is one common objective of many applications of deep learning. More concretely, forward propagators such as numerical methods can be regarded as a conditional image-to-image translation task~\cite{image2image}. Some notable works include \cite{unet,latent-diffusion,cyclegan}. Of concern is ResNet \cite{resnet} and U-Net \cite{unet}. The former adds a \textit{residual connection} which makes the model predict the shift from the input instead of the output directly, which empirically improves the performance and stability of image processing. U-Net shrinks the hidden representation in the middle of the ResNet, reducing the number of parameters and improving the globality of feature dependencies.


\subsection{Neural Operators for Solving PDEs}

There have been a great number of research works on applying neural networks to solve PDEs. These works can be largely classified into two categories: (1) approximating the solution function and (2) approximating the solution operator.

The former category is pioneered by physics-informed neural networks (PINNs) \cite{pinn}, a deep learning framework for solving PDEs in physics. The framework uses an FFN to approximate the solution to PDEs by learning the distribution of training data while minimizing the loss function that enforces constraints based on physics laws. A series of improvements to PINNs have been proposed. These include dividing the solution domain to speed up the convergence \cite{cpinn,xpinn,apinn}, combining the numerical derivative and adaptive derivative reverse propagation to improve accuracy \cite{can-pinn}. Some works focus on improving the neural architecture \cite{gapinn}, by adopting convolutional layers instead of fully connected layers, such as PhyGeoNet \cite{phygeonet}, PhyCRNet \cite{phycrnet}, etc. However, these methods have limited applicability and only a few are evaluated on complex flow equations. Moreover, since PINNs approximate one solution function, they have to be retrained for every new input function or condition.

The second category learns a whole family of solutions by learning the mapping from input functions to output functions. \cite{pino} have proved that the neural operator can approximate any arbitrary nonlinear equations. Some notable neural operators include FNO \cite{fno}, LNO \cite{lno}, and KNO \cite{kno}, among others. These operators are forward propagators similar to numerical methods but learn in other domains to achieve mesh independence. Another series of neural operators is the DeepONet \cite{deeponet}, which encodes the query location and the input functions independently and aggregates them to produce the prediction at the query location. Many improvements based on DeepONet have been proposed \cite{pino,bayesian-deeponet,SVD-and-deeponet,long-time-prediction-deeponet,shift-deeponet, physics-informed-deeponet,improved-deeponet}. Because the number of existing neural operators is too great for us to evaluate every single one, we have selected a few representative ones in this paper to include as baselines. We encourage future works to evaluate on CFDBench to compare against these baselines.

\subsection{Benchmarking Data-Driven Scientific Modeling}

Following the popularity of data-driven methods in scientific models with the help of deep learning, several evaluation benchmarks have been proposed. WeatherBench \cite{weatherbench} is a benchmark consisting of historical weather data extracted and processed from the ERA5 archive, but weather forecasting is one specific CFD problem where the geometry never changes, and there is large-scale historical data, which does not exist for many other scenarios. PDEBench \cite{pdebench} is a multi-task benchmark on several classic PDE problems, but it only includes two CFD problems where the BCs are always periodic, and physical properties and geometries are also constant. MegaFlow2D \cite{megaflow-2d} is a large-scale CFD-specific benchmark with 2 million snapshots of generated flows. However, this benchmark contains only two problems, and the BCs and physical properties are the same across all snapshots, rendering it unable to measure neural operators' generalization along these dimensions. Table \ref{tab:benchmarks-comparison} presents a comparison between the main benchmarks for data-driven PDE modeling and our benchmark.

\begin{table}[H]
    \centering
    \begin{tabular}{l|c|ccc}
        \toprule
            &   &  \multicolumn{3}{c}{Varying...} \\ 
        Name    
            & \# Problems 
            & BCs 
            & Physical Properties 
            & Geometry \\
        \midrule
        WeatherBench \cite{weatherbench}    & 1  & \xmark & \xmark & \xmark \\
        PDEBench \cite{pdebench}        & 11 & \xmark & \xmark & \xmark \\
        MegaFlow2D \cite{megaflow-2d}    & 2  & \xmark & \xmark & \cmark \\
        CFDBench (ours) & 4  & \cmark & \cmark & \cmark \\
    \bottomrule
    \end{tabular}
    \caption{A comparison between CFDBench (ours) and existing benchmarks in data-driven PDE modeling. ``Varying x'' means that different data examples have different x.}
    \label{tab:benchmarks-comparison}
\end{table}

\section{CFDBench}

\begin{figure}[H]
    \centering
    \includegraphics[width=\textwidth]{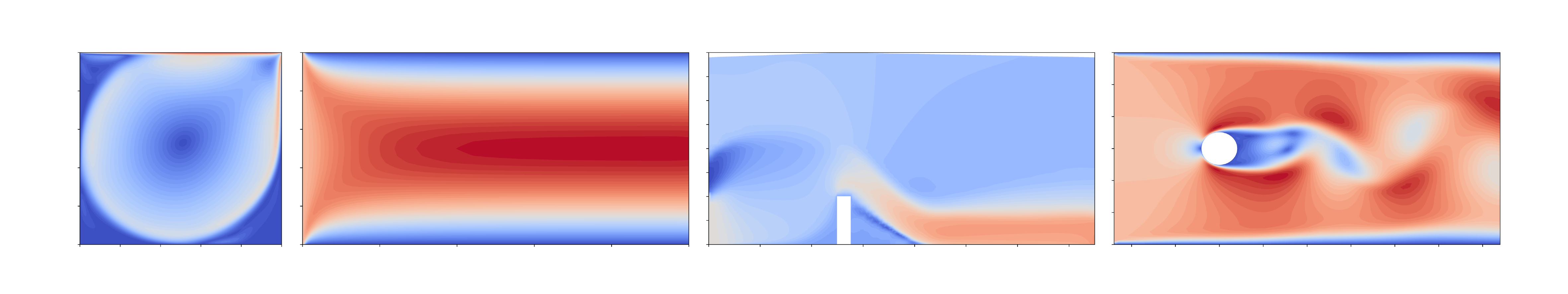}
    \caption{Some examples of the velocity field in the four problems in CFDBench. From left to right: cavity flow, tube flow, dam flow, and cylinder flow.}
    \label{fig:fluid-examples}
\end{figure}

In this section, we first give a formal definition of the flow problems. Then, we present the four flow problems included in CFDBench, along with the parameters and various considerations during dataset construction. For each problem, we generate flows with different \textit{operating parameters}, which is the term we use to refer to the combination of the three kinds of condition: (1) the BCs, (2) the fluid physical properties (PROP), and (3) the geometry of the field (GEO). Each kind of operating parameter corresponds to one subset. In each subset, the corresponding operating conditions are varied while other parameters remain constant. The goal is to evaluate the ability of the data-driven deep learning methods to generalize to unseen operating conditions. Figure~\ref{fig:fluid-examples} shows an example snapshot of each of the problems in our dataset.

\subsection{The Definition of Flow Problems}

The Navier-Stokes equations can be formalized as follows.

\begin{align}
    \begin{cases}
        \nabla \cdot (\rho \mathbf{u})= 0\\
        \frac{\partial}{\partial t} (\rho \mathbf{u}) + \nabla \cdot (\rho \mathbf{u} \mathbf{u}) = -\nabla p + \nabla \cdot 
         \mu [\nabla \mathbf{u} + (\nabla \mathbf{u})^\top],
    \end{cases}
\end{align}

where $\rho$ is the density and $\mu$ is the dynamic viscosity, $\mathbf{u} = (u, v) ^\top$ is the velocity field, and $p$ is the pressure.

Suppose the fluid is incompressible ($\rho = const$) and the fluid is a Newtonian fluid ($\tau = \mu \frac{du}{dy}$). Combining the continuum hypothesis and Stokes' law, we get the following equations inside the flow domain (when $(x,y,t) \in \mathcal D$).

\begin{align}
    \large
    \begin{cases}
        \frac{\partial u}{\partial x} + \frac{\partial v}{\partial y} = 0 \\
        \frac{\partial u}{\partial t}+ u \frac{\partial u}{\partial x} +  v \frac{\partial u}{\partial y} 
        = - \frac{1}{\rho} \frac{\partial p}{\partial x} + \frac{\mu}{\rho} (\frac{\partial^2 u}{\partial x^2} + \frac{\partial^2 u}{\partial y^2}) \\
        \frac{\partial v}{\partial t}+ u \frac{\partial v}{\partial x} +  v \frac{\partial v}{\partial y} 
        = - \frac{1}{\rho} \frac{\partial p}{\partial y} + \frac{\mu}{\rho} (\frac{\partial^2 v}{\partial x^2} + \frac{\partial^2 v}{\partial y^2}),
    \end{cases}
\end{align}

and $(u,v)$ are constant on the boundaries ($\partial \mathcal D$).

In this work, we consider four important and representative fluid problems that can comprehensively evaluate different methods' capabilities in different problems. They are (1)~the flow in the lid-driven cavity, (2)~the flow into the circular tube, (3)~the flow in the breaking dam, and (4)~the flow around the cylinder. These flow problems cover most of the common flow phenomena. They have both open and closed systems and vary in shape. The system boundaries include both moving/stationary boundaries and velocity/pressure inlet and outlet boundaries. They include vertical flows within gravity and plane flows without gravity. Their flow characteristics include the formation of a viscous boundary layer, the formation and shedding of vortexes, and the formation of jets. They have both single-phase flow and two-phase flow, both laminar flow and turbulent flow. However, in order to ensure the cleanliness of the data, that is, to ensure that the data fully satisfy the above equation, we regard the flow as the flow of incompressible Newtonian flow, ignoring the mass transfer at the two-phase interface and the energy dissipation during the flow process.

For simplicity, we will refer to the four problems as \textbf{(1)~cavity flow}, \textbf{(2)~tube flow}, \textbf{(3)~dam flow}, and \textbf{(4)~cylinder flow}.
For each problem, we use different operating parameters and generate the flow fields using numerical methods.

\subsection{Cavity Flow}

\begin{table}
    \caption{Operating parameters of the subset in the cavity flow problem.}
    \centering
    \begin{tabular}{c|p{10cm}}
        \toprule
        \multicolumn{2}{c}{
            \includegraphics[width=0.4\linewidth]{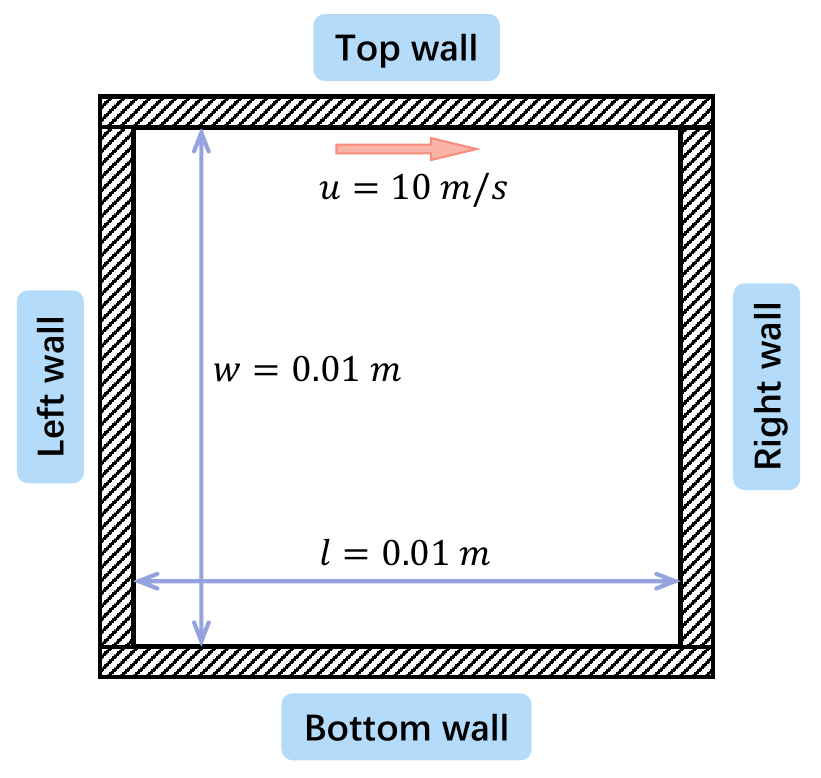}
        } \\
        \midrule
        BC 
            & $u_{\mathcal{B}} \in \{ 1, 2, 3, \cdots, 50\} m/s$ \\
        \midrule
        Property 
            & $\rho \in \{0.1, 0.5, 1, 2, 3, \cdots, 10 \} kg/m^3$ \\ 
            & $\mu \in \{ 10^{-5}, 5\times{10}^{-5},\cdots, 5\times{10}^{-3},\ {10}^{-2} \} Pa \cdot s$ \\
        \midrule
        Geometry 
            & $l,w \in \{0.01, 0.02, 0.03, 0.04, 0.05 \}$ m\\
        \bottomrule
        
    \end{tabular}
    \label{tab:dataset-composition-cavity}
\end{table}

Cavity flow refers to a flow in a square container with a moving upper wall surface (i.e., the lid) and three stationary walls. Due to viscosity, the moving wall drives the fluid in proximity to move in the same direction until the stationary wall forms a jet impacting the lower wall and then forms a secondary vortex. On the one hand, the lid-driven cavity flow has a wide range of applications in the industry, such as the transient coating (short dwell coating) process \citep{cavity-background-coating}, the ocean flow affected by the wind, and so on. On the other hand, the special case is that the BC is discontinuous \citep{cavity-background-boundary} at the connection of the moving wall and the stationary side wall, which makes it judge the convergence of numerical methods. Thus, it is widely used to verify the accuracy of computational fluid mechanics software or numerical methods \citep{cavity-background-verify}. Therefore, the construction of the top lid-driven cavity flow data set is beneficial to study the ability of the neural network model to solve the flow problem.

In the dataset with the cavity flow, the baseline conditions are $\rho = 1 kg/m^3$, $\mu = 10^{-5} Pa \cdot s$, $l = d = 0.01m$, $u_\text{top} = 10 m/s$, where $\rho$ and $\mu$ are the density and viscosity of the fluid, $l$ and $d$ are the length and width of the cavity, and $u_\text{top}$ is the top wall movement velocity. 50 different cases are generated by varying $u_\text{top}$ from $1m/s$ to $50m/s$ with a constant step size. 84 cases are generated varying the physical properties of the working fluid, with 12 different values of density and 7 values of viscosity. For the cases with different geometries, we choose different combinations of length and width from $ \{ 0.01,0.02,0.03,0.04,0.05 \} $. To have an appropriate scale of difference between the frames, we set the time step size to $ \Delta t=0.1s $.

\subsection{Tube Flow}

The tube flow refers to a water-air two-phase flow into the circular tube filling with air The boundary layer in the circular tube is one of the most common flows, which means that the viscosity resistance of the fluid on the near-wall surface is greater than the fluid in the bulk flow region. When the water flows into the round tube filled with air, we can clearly see that the flow is slow near the wall and fast in the center. Therefore, the construction of water-air laminar flow in a circular tube is beneficial to study the ability of the neural network structure to capture the two-phase interface and to learn the laminar boundary layer theory.

In the dataset of the tube flow, the baseline conditions are $\rho 
= 100 kg/m^3$, $\mu = 0.1Pa \cdot s$, $u_{in} = 1m/s$, $d = 0.1m$, $l = 1m$, where $\rho$ and $\mu$ are the density and viscosity of the fluid, $u_{in}$ is the inlet velocity (from the left), $d$ and $l$ is the diameter and the length of the circular tube. 50 cases were generated for different BCs, increasing the inlet velocity from $0.1m/s$ to $5m/s$, with increments of $0.1m/s$. 100 cases with different physical properties of the working fluid are generated, and the two-dimensional space of different densities and dynamic viscosity are shown in Table 3, where the density increases from $10kg/m^3$  to $1000kg/m^3$ with increments of $110kg/m^3$, and viscosity increases from to and the viscosity increases from $0.01 Pa \cdot s$ to $1Pa \cdot s$ with increments of $0.11 Pa \cdot s$. For different geometries, the diameter of the circular tube is taken from $\{ 0.01,0.05,0.1,0.3,0.5 \}$, and we choose five different ratios of diameter and length by making sure the length satisfies $0.1 \leq l \leq 10$.  This results in 25 different geometries. To have an appropriate scale of difference between the frames, we set the time step size to $\Delta t=0.01s$.

\begin{table}[H]
    \caption{Operating parameters of the subset in the tube flow problem.}
    \centering
    \begin{tabular}{c|p{10cm}}
        \toprule
        \multicolumn{2}{c}{
            \includegraphics[width=0.75\linewidth]{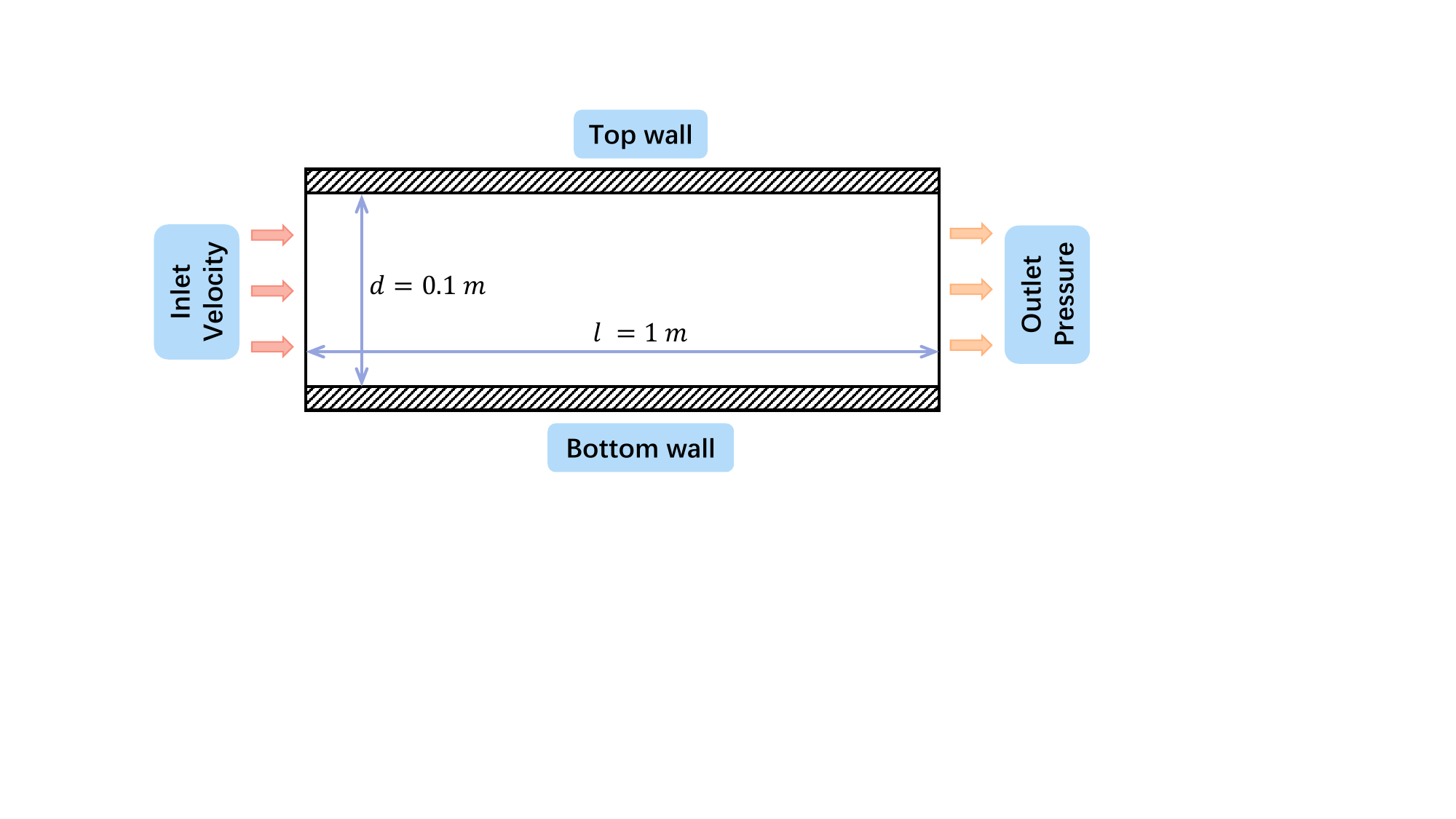}
        } \\
        \midrule
        BC
            & $u_{\mathcal{B}} \in \{0.1, 0.2, 0.3, \cdots, 5\} m/s$ \\
        \midrule
        Property 
            & $\rho \in \{10, 120, 320,\cdots, 1000\} kg/m^3$ \\
            & $\mu \in \{0.01,\ 0.12,\ 0.23,\cdots, 1\} Pa \cdot s$ \\
        \midrule
        Geometry 
            & $l \in \{0.01, 0.05, 0.1, 0.3, 0.5 \} m$ \\
            & $d/l \in \{1, 2, 5, 7.5, 10, 15, 20, 50, 75, 100\} $ \\
        \bottomrule
    \end{tabular}
    \label{tab:dataset-composition-tube}
\end{table}

\subsection{Dam Flow}

A dam is a barrier across flowing water that obstructs, directs, or slows down the flow. Meanwhile, sudden, rapid, and uncontrolled release of impounded water quickly causes a dam to burst \citep{dam-background}. To further understand the flow of water over the dam, we simplified it to the flow of water over a vertical obstacle. When the Reynolds number is low, the fluid is dominated by the viscous force and will flow vertically down the wall as it flows through the dam \citep{dam-numerical}. As the speed increases, the fluid is more affected by the inertial force, and a jet will be formed. Then the fluid falls to the boundary because of gravity and the collision with the boundary makes more reverse flow, which will hit the dam with a bigger velocity than the inlet. Therefore, the dam flow dataset is helpful in studying the learning ability of the model for flows subject to different viscous and inertial forces.

In the dataset of dam flow, the baseline conditions are $\rho 
= 100 kg/m^3$, $\mu = 0.1Pa \cdot s$, $u_{in} = 1m/s$, $h = 0.1m$, $w = 0.05m$, where $\rho$ and $\mu$ are the density and viscosity of the fluid, $u_{in}$ is the inlet velocity (from the left), $h$ and $w$ is the height and the width of the dam obstacle. The entire fluid domain is 1.5m long and 0.4m high. The inlet velocity boundary is close to the ground, with a total length of 0.1m, and 0.3m above it is the inlet pressure boundary. The barrier is located 0.5m from the entrance. 70 cases were generated for different BCs, increasing the inlet velocity from $0.05m/s$ to $1m/s$ with increments of $0.05m/s$ and from $1m/s$ to $2m/s$ with increments of $0.02m/s$. 100 cases with different physical properties of the working fluid are generated, and the two-dimensional space of different densities and dynamic viscosity are shown in Table 3, where the density increases from $10kg/m^3$  to $1000kg/m^3$ with increments of $110kg/m^3$, and viscosity increases from to and the viscosity increases from $0.01 Pa \cdot s$ to $1Pa \cdot s$ with increments of $0.11 Pa \cdot s$. 50 cases with different geometries are generated, increasing the height from $0.11m$ to $0.15m$ with increments of $0.01m$ and width from $0.01m$ to $0.09m$ with increments of $0.01m$ of dam obstacle. To have an appropriate scale of difference between the frames, we set the time step size to $\Delta t=0.1s$.

\begin{table}[H]
    \caption{Operating parameters of the subset in the dam flow problem.}
    \centering
    \begin{tabular}{c|p{10cm}}
        \toprule
        \multicolumn{2}{c}{
            \includegraphics[width=0.8\linewidth]{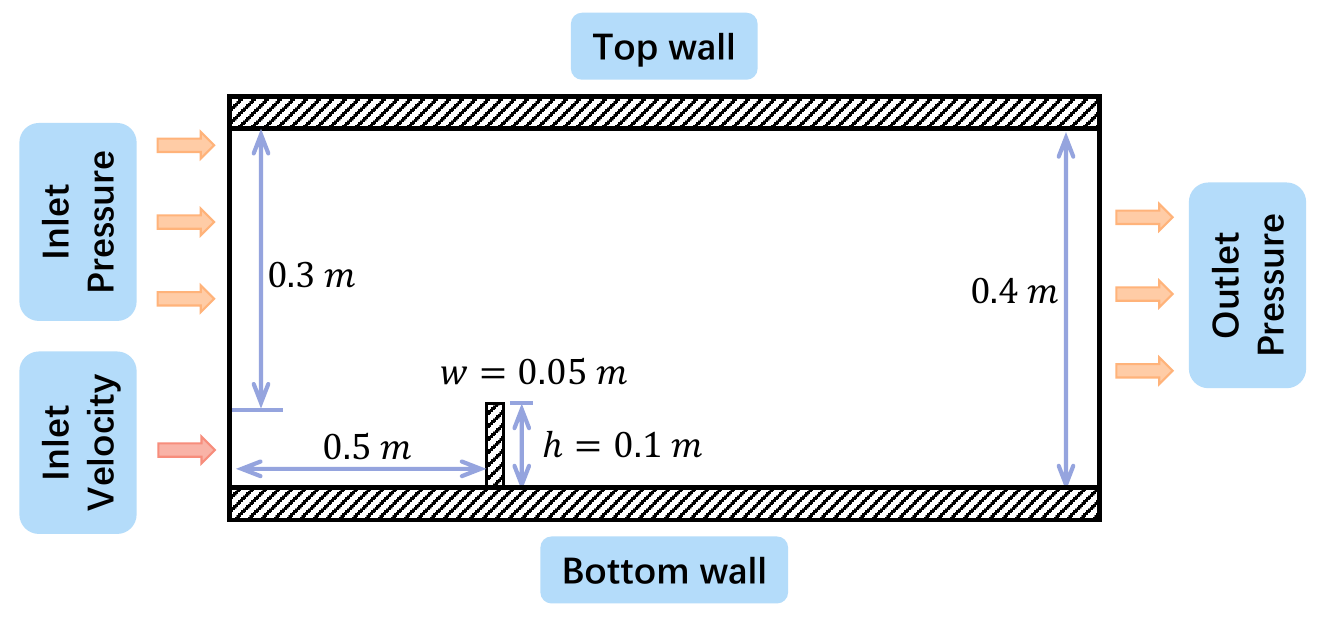}
        } \\
        \midrule
        BC 
            & $u_{\mathcal{B}} \in \{ 0.05, 0.1, \cdots, 1\} \cup \{1.02, 1.04, \cdots 2 \} m/s$ \\
        \midrule
        Property 
            & $\rho \in \{0.1, 0.5, 1, 2, 3, \cdots, 10 \} kg/m^3$ \\ 
            & $\mu \in \{ 10^{-5}, 5\times{10}^{-5},\cdots, 5\times{10}^{-3},\ {10}^{-2} \}  Pa \cdot s$ \\
        \midrule
        Geometry
            & $h \in \{0.11, 0.12, 0.13, 0.14, 0.15 \} m$ \\
            & $w \in \{0.01, 0.02, \cdots, 0.08, 0.09 \} m$ \\
        \bottomrule
    \end{tabular}
    \label{tab:dataset-composition-dam}
\end{table}

\subsection{Cylinder Flow}

A flow around a cylinder is a typical boundary layer flow, which is commonly seen in the industry where water flows through bridges, the wind blows through towers, etc \citep{cylinder-background-boundary}. When the fluid with a large flow rate passes around the cylinder, the boundary layer fluid separates to form the reverse zone due to the combined effect of reverse pressure gradient and wall viscous force retardation. At a specific Reynolds number, the two sides of the cylinder periodically generate a double row of vortexes with opposite rotational directions and are arranged in a regular pattern. Through nonlinear interactions, these vortexes form a Karman vortex street. after nonlinear action. Therefore, the cylindrical flow dataset is important for examining the capability of neural networks in modeling periodic flows with obstacles.

In the dataset of the cylinder flow, the baseline conditions are $\rho=10kg/m^3$, $\mu=0.001Pa\cdot s$, $u_{in}=1 m/s$, $d=0.02m$, $x_1=y_1=y_2=0.06m$, $x_2=0.16m$, where $\rho$ and $\mu$ are the density and viscosity of the fluid, $u_{in}$ is the inlet velocity (from the left), d is the diameter of the cylinder, $x_1,x_2,y_1,y_2$ is the distance between the center of the cylinder and the left, right, top and bottom boundaries, respectively. 50 cases are generated for different BCs, increasing the inlet speed from $0.1m/s$ to $5m/s$ with increments of $0.1m/s$. 115 cases are generated for the different physical properties of the fluid so that the Reynolds numbers are in the range of $[20,1000]$. Table 4 shows some values of density and viscosity, but not all combinations are used because that results in Reynolds numbers outside of the target range. For different geometries, the distance from the cylinder to the upper and lower boundaries and the entrance is taken from $\{ 0.02,0.04,0.06,0.08,0.1 \}$, the distance from the cylinder to the exit boundary is taken from $\{0.12,0.14,0,16,0.18,0.2\}$, and the radius of the cylinder is taken from $\{0.01,0.02,0.03,0.04,0.05\}$. 20 cases are generated. To ensure an appropriate scale of difference between the frames, we set the time step size to $\Delta t=0.001s$.

\begin{table}[H]
    \caption{Operating parameters of the subset in the cylinder flow problem.}
    \centering
    \begin{tabular}{c|p{10cm}}
        \toprule
        \multicolumn{2}{c}{
            \includegraphics[width=0.8\linewidth]{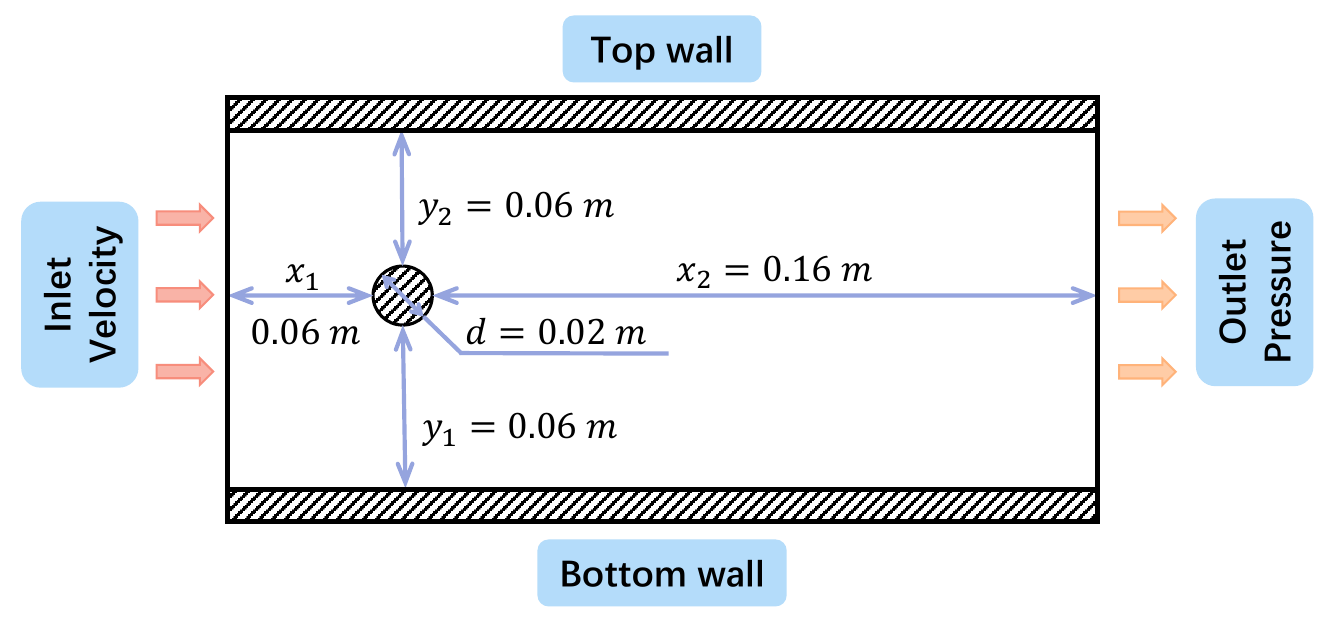}
        } \\
        \midrule
        BC 
            & $u_{\mathcal{B}} \in \{0.1, 0.2, 0.3, \cdots, 5 \} m/s$ \\
        \midrule
        Property 
            & $\rho \in \{0.1, 0.2, \cdots, 1 \} \cup \{1.5, 2.5, \cdots, 4.5,  5 \} \cup \{  6,   7, \cdots,   9, 10 \}$ \\
            & $\cup \{ 20,  30, 40, \cdots, 250 \}\cup \{300, 400, 500\}  kg/m^3 $ \\
            & $\mu \in \left \{ 10^{-4}, 5 \times 10^{-4}, 10^{-3}, 5 \times 10^{-3}, 10^{-2} \right \}  Pa \cdot s $ \\
        \midrule
        Geometry 
            & $d \in \{0.01, 0.02, 0.03, 0.04, 0.05 \} m$ \\
            & $x_1, y_1, y_2 \in \{0.02, 0.04, 0.06, 0.08, 0.1\} m$ \\
            & $x_2 \in \{0.12, 0.14, 0.16, 0.18, 0.2 \} m$ \\
        \bottomrule
    \end{tabular}
    \label{tab:dataset-composition-cylinder}
\end{table}

The problem datasets and the number of cases under different operating conditions are summarized in Table~\ref{tab:dataset-breakdown}.

\begin{table}[H]
    \centering
    \caption{Breakdown of the number of cases in each problem (the rows) and the corresponding subsets (the columns) in CFDBench. Each problem contains three subsets, each with one type of operating condition parameter that is varied.}
    \begin{tabular}{c|ccc|c|c|cc}
        \toprule
            & \multicolumn{4}{c|}{\textbf{Number of cases}} 
            & 
            & \multicolumn{2}{c}{Each frame}\\
        \midrule
        \textbf{Problem} 
            & \textbf{BC} 
            & \textbf{PROP} 
            & \textbf{GEO} 
            & \textbf{Total} 
            & \textbf{\# frames} 
            & \textbf{File Size}
            & \textbf{Gen. Time} 
        \\
        \midrule
        Cavity Flow
            & 50    & 84   & 25     & 159 & 34,582  & 5.2 MB & 0.92s \\ 
        Tube Flow
            & 50    & 100  & 25     & 175 & 39,553  & 4.8 MB & 1.08s \\ 
        Dam Flow
            & 70    & 100  & 50     & 220 & 21,916  & 2.0 MB & 3.98s \\ 
        Cylinder Flow
            & 50    & 115  & 20     & 185 & 205,620 & 4.4 MB & 1.18s \\ 
        \midrule
        Sum 
            & 220   & 399  & 120    & 739 &   301,671  & \\
        \bottomrule
    \end{tabular}
    \label{tab:dataset-breakdown}
\end{table}

\subsection{Data Generation}

All the data in this paper are generated by ANSYS Fluent 2021R1. In order to calculate the viscosity term accurately, the laminar model is used for laminar flow and SST $k- \omega$ model for turbulent flow. All solvers used are based on pressure. We choose a Coupled Scheme for single-phase flow and SIMPLE for two-phase flow as a pressure-velocity coupling algorithm. The pressure equation uses the second-order interpolation method (the VOF model uses the PRESTO! Interpolation method), and the momentum equation adopts the second-order upwind method. The time term adopts the first-order implicit format and interpolation uses the least squares method. To capture the phenomenon of boundary layer separation at the near-wall surface, the size of the first layer mesh in the near-wall surface is encrypted to $10^{-5} m$. To ensure the accuracy of the computational model and results, all computational models underwent grid-independent validation. 

After discretizing the governing equations, the conservation equation of the universal variable$(\Phi_P)$ at the grid element $P$ can be expressed as:

\begin{equation}
    a_P \Phi_P = \sum_{nb} a_{nb} \Phi_{nb} + b
\end{equation}

in which $a_P$ is coefficient of the node of element $P$, $a_{nb}$ is coefficients of neighbor nodes and $b$ is the coefficient generated by constant term, source term and boundary condition.
It defines the global scaling residual as:

\begin{equation}
   R^{\Phi} = \frac{\sum_{cells} |\sum_{nb} a_{nb} \Phi_{nb} + b - a_P \Phi_P|}{\sum_{cells} |a_P \Phi_P|} 
\end{equation}

\begin{figure}[H]
    \centering
    \includegraphics[width=0.9\linewidth]{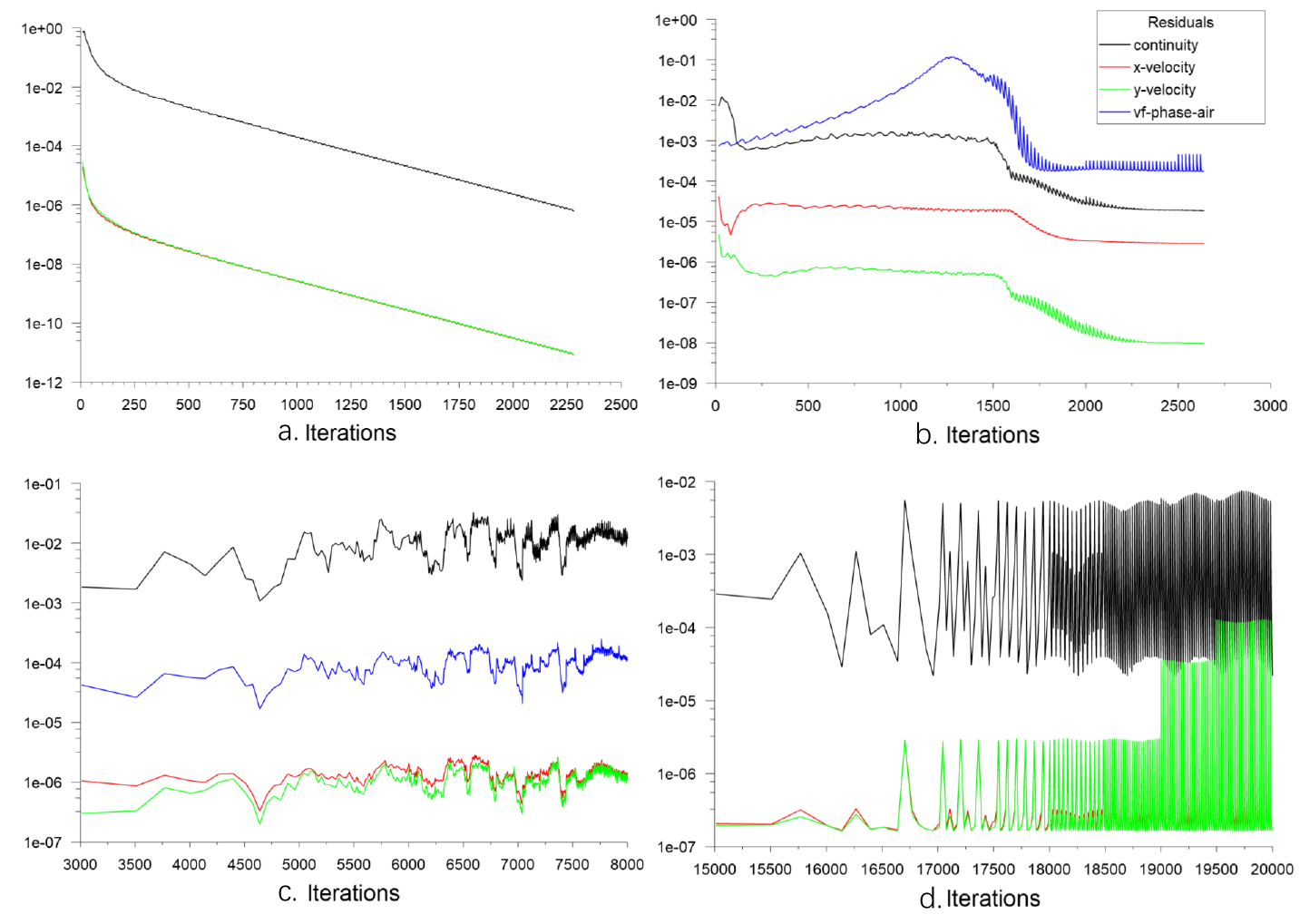}
    \caption{The residuals of each flow problems in this paper. (a) cavity flow, (b) tube flow, (c) dam flow, (d) cylinder flow.}
    \label{fig:residuals}
\end{figure}

The residual represents the relative size of the total unbalance term in the computational domain, and is generally used to judge the convergence of the solution. The smaller the residual, the better the convergence. In this paper, the residual convergence condition of all terms is set to $10^{-9}$, and the residuals in the final calculation results are shown as Figure~\ref{fig:residuals}. The residuals of the velocity terms are all at least $10^{-6}$. 

All generations are run with 30 solver processes on a CPU of AMD Ryzen Threadripper 3990X. The final generated data was interpolated to a grid size of $64 \times 64$.

\subsubsection{Training Subsets}

We divide the data of each problem into three subsets, BC, PROP, and GEO. Then, we use the seven different combinations (PROP, BC, GEO, PROP + BC, PROP + GEO, BC + GEO, All) to evaluate the generalization ability along different dimensions of various baseline methods. The specific number of data in the subsets is shown in ~\ref{tab:dataset-breakdown}, and each combination is the sum of all the data of the two subsets.

\subsubsection{Data Splitting}

Each subset of data is split into training, validation, and test sets with a ratio of 8:1:1. To ensure that the operating parameters in the test set are never seen during training, we require that the frames of each case are never distributed among different splits.

\section{Experiments}

After generating the benchmark data, we use it to train popular data-driven neural networks that can be used for approximating the solutions to PDEs. To keep the number of experiments manageable, in the following discussions, unless stated otherwise, we have the models predict the velocity field. We believe that modeling other properties or components of the flow should not be too different. 

We first define the learning objective of the neural network. Then, we give a brief description of the baselines we experimented on. After that, we explain the loss functions and hyperparameters used in the experiments.

\subsection{Training Objectives}

Most flow problems focus on solving the distribution of flow fields in the domain. Therefore, the objective of the neural networks is to approximate the following mapping within the domain $\mathcal{D}= \{(x,y,t) \mid x \in [a,b],y \in [c,d],t \in [0, T] \}$:
 
\begin{equation}
G: (\Sigma, \Omega) \mapsto u
\end{equation}

where $\Omega = \left(u_\mathcal{B},\rho,\mu,d,l,w\right)$ is the operating parameters, which include the BC $(u_\mathcal{B})$, the physical properties $(\rho,\mu)$, and the geometry $S$. $\Sigma$ is the input function, which can be either the velocity field at a certain time (in autoregressive generation) or the spatiotemporal coordinate vector $(x,y,t)$ (in the non-autoregressive model). $u$ is the output function, which is the velocity field.

When using a neural network $f_\theta$ with parameters $\theta$ to approximate $G$, there are two approaches: non-autoregressive and autoregressive modeling.

\subsubsection{Non-Autoregressive Modeling}

In non-autoregressive modeling, the input function $\Sigma$ is a \textit{query location} $\left(x,y,t\right)$ and the model directly outputs the solution at that position:

\begin{equation}
\hat{u}\left(x,y,t\right)=f_\theta((x,y,t),\ \Omega) \in \mathbb R
\end{equation}

\subsubsection{Autoregressive Modeling}

Autoregressive modeling, which is similar to traditional numerical methods, learns the mapping of a flow field from the current time step to the next time step. Therefore, it predicts the distribution of flow fields at each moment according to the temporal order:

\begin{equation}
\hat{u} \left(t\right) = f_\theta \left( u \left( t - \Delta t \right), \Omega \right) \in \mathbb R ^ {n \times m}
\end{equation}

where $\hat{u}$ is the predicted value at time $t$, $n$ and $m$ are the height and width of the domain. In other words, the input function is $\Sigma = u \left(t - \Delta t \right)$.

The learning goal is to find one $\theta^\ast$ that minimizes the loss function $\mathcal{L}$ on the training data $\mathcal T$.
 
\begin{equation}
    \theta^\ast = \arg \max_{\theta}{\mathcal{L}(u\left(x,y,t\right),\hat{u}(x,y,t))}  
        \quad \forall x,y,t\in\ \mathcal{D}, u \in \mathcal T
\end{equation}

\subsection{Baselines}

We evaluate on CFDBench some popular and performant neural networks that have been applied to solve PDE in existing works. Although CFDBench can be used to evaluate both data-driven and physics-informed methods, our experiments are limited to the former. This is because most physics-informed methods enforce operating conditions through loss functions, requiring retraining on unseen conditions.

We can generally predict the flow in two manners: non-autoregressively or autoregressively. The former directly predicts the output function value at a query location specified in the input. The latter predicts the field at the next time step given the field at the current time step. The two kinds are not directly comparable, so we discuss them separately. 

\begin{figure}[!t]
    \centering
    \includegraphics[width=0.95\linewidth]{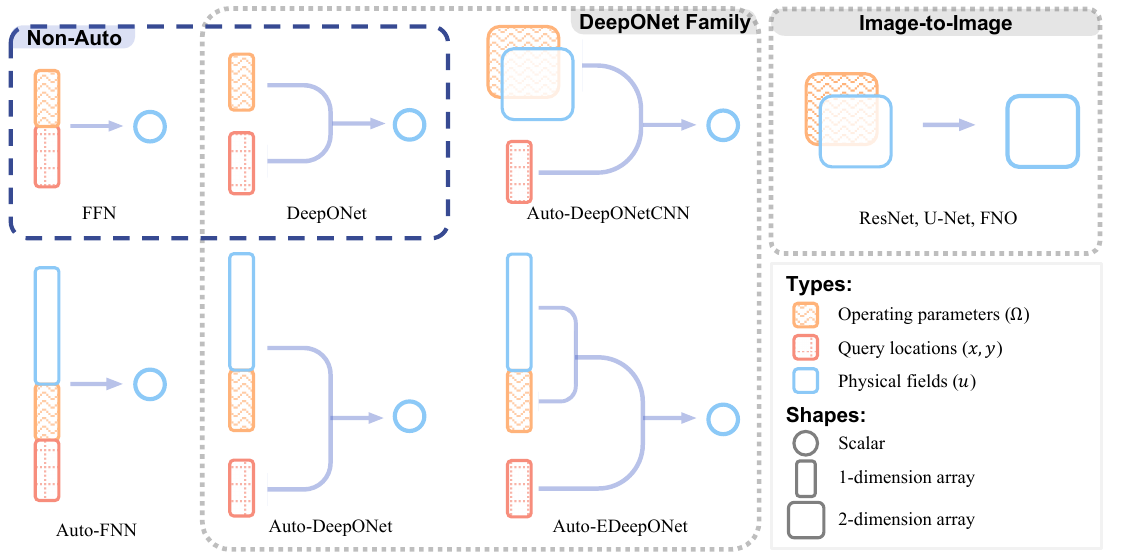}
    \caption{Overview of the input and output types and shapes of each baseline model.}
    \label{fig:network-io}
\end{figure}

\begin{figure}[!t]
    \centering
    \includegraphics[width=0.98\linewidth]{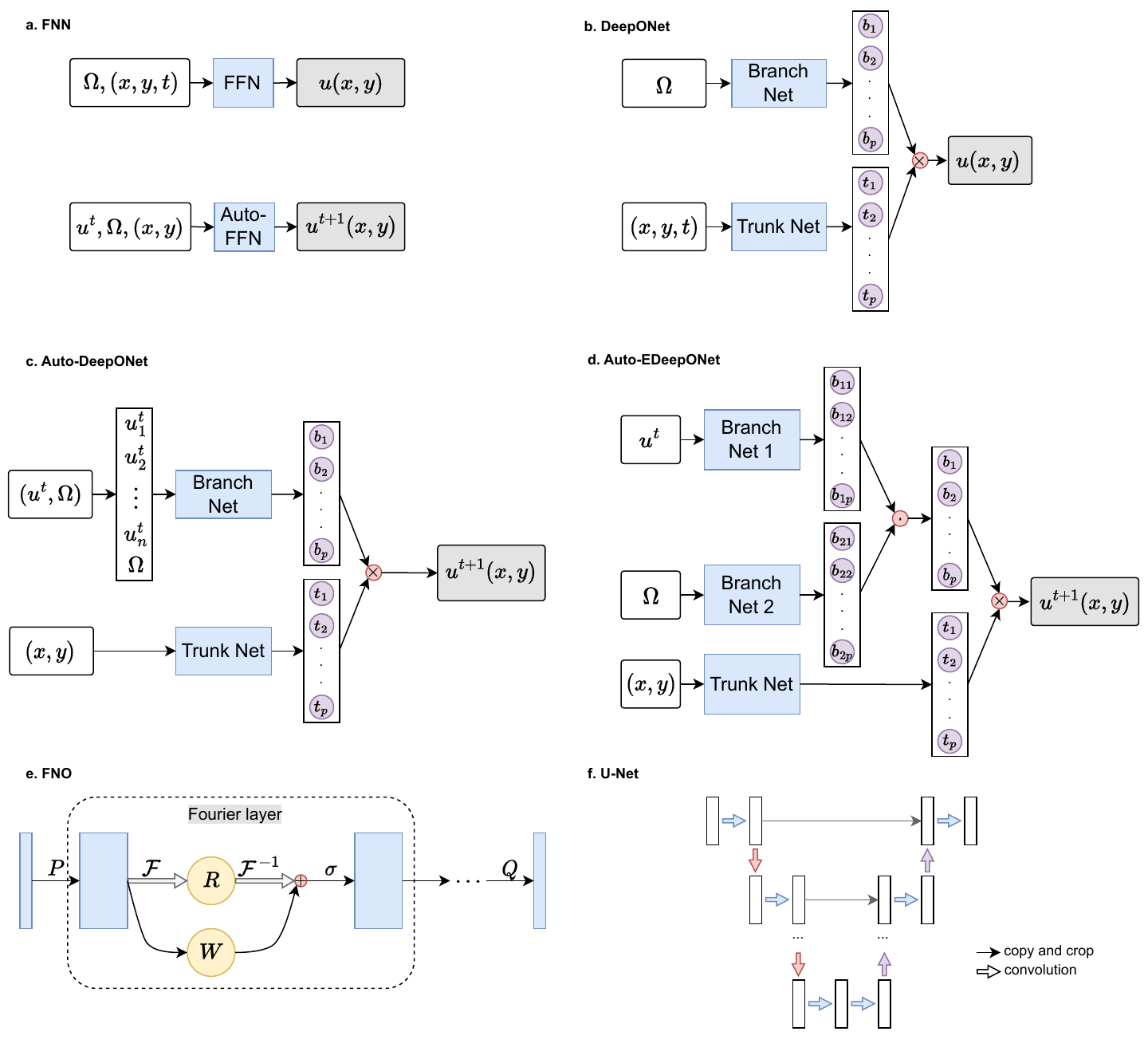}
    \caption{The structure of each baseline model in this paper.}
    \label{fig:networks-structure}
\end{figure}

From the perspective of the model architecture, we can categorize them into three types: (1) FFNs, (2) the DeepONet family, and (3) image-to-image models. Models within the first category simply concatenate all inputs into one vector and maps that to the prediction space with an FFN, shown on the left of Figure \ref{fig:network-io}. The second category includes all variants of DeepONet~\cite{deeponet}, shown in the middle of Figure \ref{fig:network-io}. The essence of this architecture is that the query location is independently encoded by a \textit{trunk net}. This makes it possible to encode the input functions and other conditions without being limited to the shape or mesh of the output function domain and reuse that encoding to query the value of the output function at any location. The third category contains ResNet, U-Net, and FNO, shown on the right of Figure \ref{fig:network-io}. They are the models that accept a $n$-dimensional array and output another $n$-dimensional array, which is the architecture that is commonly used for image-to-image tasks. Thus, we name this category image-to-image models. Table~\ref{tab:model-overview} compares all the baselines that we consider in this paper and Figure~\ref{fig:network-io} figuratively illustrates the types and shapes of the input and output of each model.\footnote{In this paper, we regard all models that accept a query location, and have an independent network (i.e., the trunk net) for encoding query locations, and predict values at those locations. However, the differences between the DeepONet family and image-to-image models are subtle and may be viewed as some variants of one another.}

\begin{table}[H]
    \caption{
        Overview of the different baseline models we consider. \textbf{``Auto.''} refers to whether the method is autoregressive. $u_\text{sample}$ is a list of samples points from $u$.
    }
        \begin{tabular}{cccccc}
            \toprule
             & & \multicolumn{2}{c}{\textbf{Shape}} \\
            \textbf{Method}     
                & \textbf{Auto.} 
                & \textbf{Input}
                & \textbf{Output}
                & \textbf{Inputs}
                & \textbf{Outputs} \\
            \midrule
            FFN
                & No
                & Any
                & Any
                & $(x,y,t), \Omega$ 
                & $ \hat u(x,y,t) $ \\
            DeepONet 
                & No
                & Any
                & Any
                & $(x,y,t), \Omega$ 
                & $ \hat u(x,y,t) $ \\
            Auto-FFN 
                & Yes
                & Any 
                & Any
                & $u_\text{sample}(t - \Delta t), \Omega, (x, y)$
                & $ \hat u(x, y, t) $ \\
            Auto-DeepONet 
                & Yes
                & Any
                & Any
                & $u_\text{sample}(t - \Delta t), \Omega, (x, y)$
                & $ \hat u(x, y, t) $\\
            Auto-EDeepONet
                & Yes
                & Any
                & Any
                & $u_\text{sample}(t - \Delta t), \Omega, (x, y)$
                & $ \hat u(x, y, t) $ \\
            Auto-DeepONetCNN
                & Yes
                & Grid 
                & Any 
                & $u(t - \Delta t), \Omega, (x, y)$
                & $ \hat u(x, y, t) $ \\
            ResNet
                & Yes
                & Grid 
                & Grid
                & $u(t-\Delta t), \Omega$
                & $ \hat u(t) $\\
            U-Net 
                & Yes
                & Grid 
                & Grid
                & $u(t-\Delta t), \Omega$
                & $ \hat u(t) $ \\
            FNO 
                & Yes 
                & Grid 
                & Grid
                & $u(t-\Delta t), \Omega$
                & $ \hat u(t) $ \\
            \bottomrule
        \end{tabular}
    \label{tab:model-overview}
\end{table}

In \ref{sec:baseline}, we will briefly describe the structure of each baseline model.

\subsection{Conditioning on Operating Parameters}

Most existing works on neural operators keep the operating parameters ($\Omega$) constant, and the input function, which is the IC, is the only input to the operator. In contrast, CFDBench considers varying the operating parameters while keeping the IC constant. Consequently, we need to make appropriate modifications to existing neural models for PDEs such that the predictions can be conditioned on the operating parameters.

For the autoregressive models, we treat the problem as a conditional image-to-image translation task, where the velocity field at the previous moment $u(x,y,t-\Delta t)$ is the input image, the velocity field at the current moment $u(x,y,t)$ is the target image, and the operating condition parameters $\Omega$ are the condition. For simplicity, we add $\Omega$ to the input as additional channels, one channel for each parameter. In this work, there are 5 parameters in $\Omega$, so the input at position $(x,y)$ is $(u(x,y),u_\mathcal{B},\rho,\mu,h,w)$ where $h,w$ are the height and width. For the flow around a cylinder, the model also needs to know the location and shape of the obstacle. To this end, we add a \textit{mask} channel where 0 indicates obstacles at that position and 1 indicates no obstacles.

\subsection{Loss Functions}

During training, we use the normalized mean squared error (the NMSE defined below) as the training loss function to ensure that the model would prioritize minimizing the difference for labels with smaller absolute values.\footnote{Preliminary experiments show that using a different loss function for training (e.g., using MSE instead of NMSE) does not impact the primary conclusions about the behaviors of the models that are drawn from the results. The only significant behavior change is that the loss function used for training will be smaller on the test data. Thus, this work only trains the baseline models using NMSE.} For evaluation, we also report the following three kinds of error values for comprehensiveness. We denote the label value with $\mathbf{Y}$ and the predicted value with $\hat{\mathbf Y}$.

\paragraph{Mean Square Error (MSE)}

\begin{equation}
\text{MSE} = \frac{1}{n} \sum_{i=1}^n(\mathbf Y_i-\hat{\mathbf Y_i})^2
\end{equation}

\paragraph{Normalized Mean Square Error (NMSE)}

\begin{equation}
\text{NMSE} = \frac{\sum_{i=1}^n (\mathbf Y_i-\hat{\mathbf Y_i})^2}{\sum_{i=1}^n \mathbf Y_i ^2}
\end{equation}

\paragraph{Mean Absolute Error (MAE)}

\begin{equation}
\text{MAE} = \frac{1}{n} \sum_{i=1}^n | \mathbf Y_i-\hat{\mathbf Y _ i}|
\end{equation}

As we will show with experiments in Section~\ref{sec:results}, one method may perform better than another method in terms of one metric, but perform worse in terms of another metric. Therefore, it is important for practitioners to select one or multiple metrics that best reflect their interests.

\subsection{Hyperparameter Search}

The performance of the methods is dependent on the hyperparameters such as learning rate, number of training epochs, etc.  Because our problem setting is significantly different from existing works, the optimal hyperparameters of each baseline model are likely very different from the ones found by the authors. We perform a hyperparameter search of the baseline models using the PROP subset of the cavity flow problem (84 different flows).

A more detailed description of the hyperparameter search process can be found in the Appendix. In summary, to make the methods comparable, we generally want to keep the number of parameters to be roughly the same.\footnote{An alternative for fair comparison is to align the computational cost (for training and testing) of the models, which is another important practical concern in the application of CFD modeling methods.} For ResNet, U-Net, and FNO, we try different depths and numbers of hidden channels. We also experiment with new ways to inject operating parameters. For FNN and variants of DeepONets, we try different widths and depths of the hidden linear layers. Additionally, the learning rate is selected individually for each method based on the validation loss, and we always train until convergence.
\subsubsection{ResNet}

\begin{table}[H]
    \caption{The validation loss of ResNet and the identity transformation for the 7 subsets (see Section~\ref{sec:results}) in the cavity flow problem. The better result is highlighted in \textbf{bold}.}
    \centering
    \setlength\tabcolsep{4pt}
    \begin{tabular}{c|ccccccc}
        \toprule
        Method & (1) & (2) & (3) & (4) & (5) & (6) & (7) \\
        \midrule
        \multicolumn{8}{c}{NMSE} \\
        \midrule
        Identity
            & 0.100
            & \textbf{0.108}
            & \textbf{0.076}
            & \textbf{0.108} 
            & 0.097 
            & \textbf{0.112} 
            & 0.111 
            \\
        ResNet 
            & \textbf{0.065}
            & 0.147
            & 0.863
            & 0.200
            & \textbf{0.094}
            & 0.156
            & \textbf{0.080}
            \\
        
        \midrule
        \multicolumn{8}{c}{MSE} \\
        \midrule
        
        Identity
            & 0.343
            & \textbf{0.031}
            & \textbf{0.029}
            & 0.149
            & 0.028
            & \textbf{0.347}
            & 0.164
            \\
        ResNet
            & \textbf{0.065}
            & 0.044
            & 0.500
            & \textbf{0.119}
            & \textbf{0.027}
            & 0.339
            & \textbf{0.058}
            \\
        
        \midrule
        \multicolumn{8}{c}{MAE} \\
        \midrule

        Identity
            & 0.166
            & \textbf{0.076}
            & \textbf{0.057}
            & \textbf{0.120}
            & \textbf{0.067}
            & \textbf{0.167}
            & \textbf{0.119}
        \\
        ResNet
            & \textbf{0.112}
            & 0.146
            & 0.624
            & 0.166
            & 0.098
            & 0.296
            & 0.130
        \\
        
        \bottomrule
    \end{tabular}
    \label{tab:resnet-result}
\end{table}

For ResNet, we conducted a hyperparameter search on the depth $d$ (i.e., the number of residual blocks) and hidden dimension $h$ (i.e., the number of channels of the output of each residual block). We found that ResNet’s ability to learn from flow problems is poor, and it quickly becomes unable to converge when $d$ and $h$ increase\footnote{We regard situations where the prediction is worse than the identity transformation as failure to converge.}. The setting with the lowest validation loss is $d=4$ and $h=16$, which we used to train on the data of flow in the tube, and the test loss is shown in Table~\ref{tab:resnet-result}. The result shows that ResNet’s performance is generally slightly worse than the identity transformation. One plausible explanation for this is that ResNet is poor at modeling global dependencies, i.e., the input signal at any point after one convolution layer with a $k \times k$ kernel can only spread around the original position within its neighboring $k \times k$ range. Therefore, we do not consider ResNet in further discussions below.

\subsection{Other Details}

For autoregressive models, we always train the model on one forward propagation, while for non-autoregressive models to train on randomly sampled query points on the entire spatiotemporal domain.
We tune the learning rate on the cavity PROP subset, and always have it decay by a factor of 0.9 every 20 epochs, which we empirically found to be effective. One may get better performance by tuning more hyperparameters, such as trying different learning rate schedulers and tuning them on the entire dataset. However, that is prohibitively expensive considering the size of the dataset. 

All methods were implemented using PyTorch deep learning framework, and all experiments were executed on one local computer with one RTX 3060 GPU. Most results are the average of three runs with different random seeds. 

\section{Results}
\label{sec:results}

Our analysis of the experimental results commences with the prediction of the flow field distribution at a singular time step, subsequently progressing to the autoregressive inference of multiple sequential time steps. To evaluate the predictive capabilities, we conduct a comparative assessment of both non-autoregressive and autoregressive models. Additionally, we provide a comparative analysis of the computational power consumption associated with each of these models.

\subsection{Single Step Prediction}

Figure~\ref{fig:results-velocity-1} and Figure~\ref{fig:results-velocity-2} show the predicted velocity field of all baseline models on the three subsets of the four flow problems in CFDBench. From top to bottom, the first row is the input, the second row is the label, and the following are the predictions of non-autoregressive and autoregressive models. We find, in general, that the baseline models perform relatively well on cavity flow and dam flow while struggling on tube flow and cylinder flow, especially for non-autoregressive models. 

It is important to recognize the difference between autoregressive and non-autoregressive models when analyzing the result. The task of the non-autoregressive model is to directly produce the value of the output function at a designated query location in the entire spatiotemporal domain. This should be significantly more difficult than the autoregressive model, which only needs to learn the mapping from the field at the previous time frame to the field at the current time frame. 

Also, the autoregressive models require that the input and output functions be represented with a grid, which limits their flexibility and may result in loss of information on regions where the field value changes sharply for small spatial changes. Furthermore, the non-autoregressive model has better mesh-independence, because the model can output the predicted value of the output function at any location. This has great significance for the study of many topographic complex problems. In addition, non-autoregressive inference may be much more efficient~\footnote{Efficicent in terms of inference speed, compared with autoregressive models of roughly the same size.} because it can predict values at any time frame while autoregressive models need to propagate one time step at a time. In summary, the autoregressive and non-autoregressive models cannot be directly compared against each other, and non-autoregressive inference is generally much faster at long-range prediction and is significantly more difficult.

\subsubsection{Non-Autoregressive Modeling}

\begin{figure}[!t]
    \centering
        \includegraphics[width=0.9\linewidth]{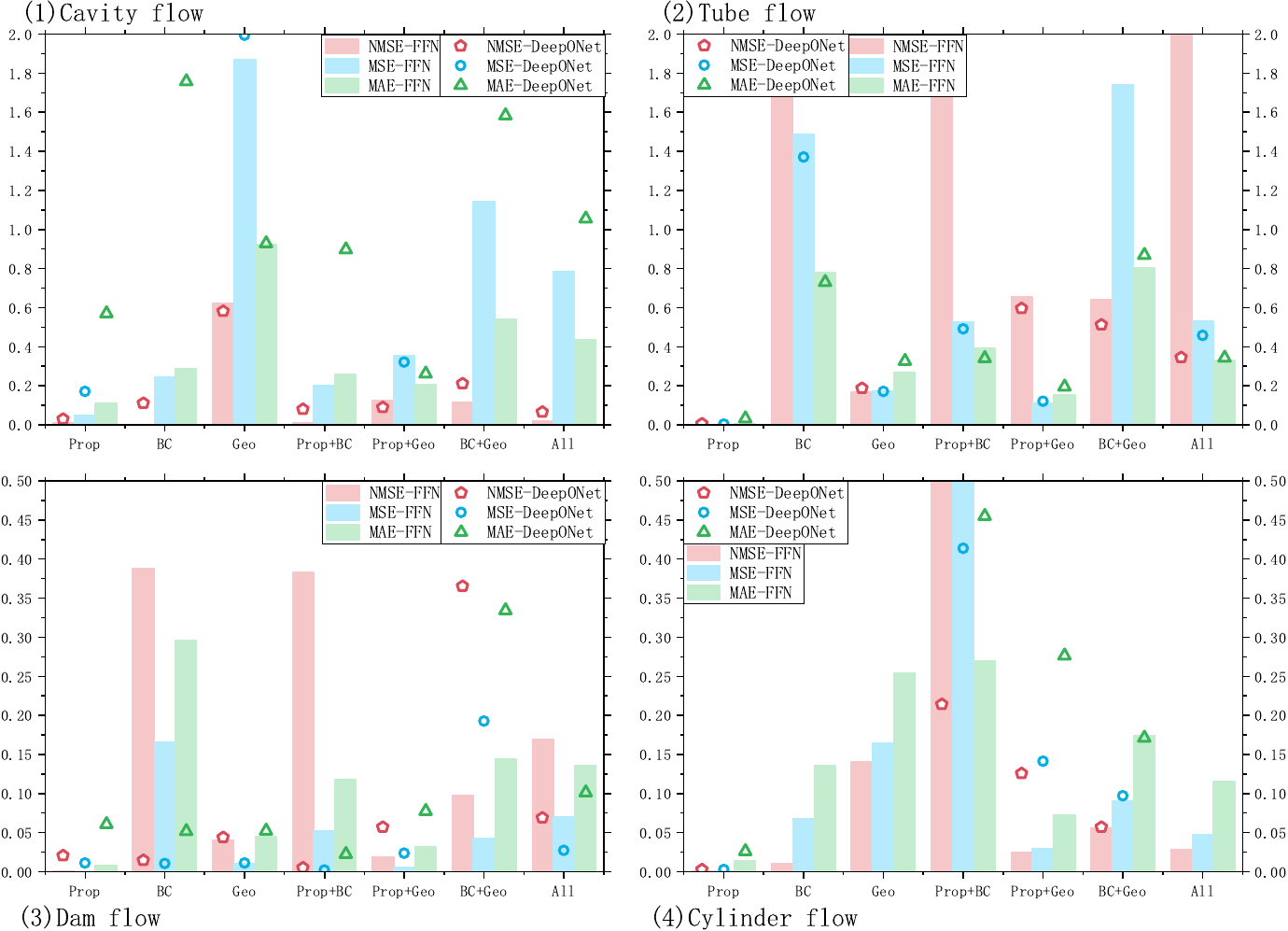}
        \label{fig:enter-label}
        \caption{The prediction results of 2 non-autoregressive baseline models on 7 subsets of the data set, with the vertical axis representing the average NMSE of all frames in the test set and the horizontal axis representing the data type.}
        \label{fig:result-none-auto}
\end{figure}

Figure ~\ref{fig:result-none-auto} shows the test NMSE, MSE, and MAE of  FFN and DeepONet on the four problems and their corresponding seven subsets in CFDBench. More specific results can be found in Table ~\ref{tab:non-auto-results}. Contrary to the observations by \citep{deeponet}, we find that that is no clear winner between FNN and DeepONet in terms of generalization ability. 
Moreover, we observe that the error is generally fairly large compared to the numerical methods that are used to generate the data, and this order of magnitude is arguably not suitable for many practical applications of fluid simulation. This indicates that pure data-driven neural operators still have a large room for improvement before they may be used to replace traditional numerical methods.

\subsubsection{Autoregressive Modeling}

We also observe the PROP subset is generally easier than other subsets. This is likely because physical properties affect the velocity less than other operating parameters, making the train-test domain gap smaller.
With varying BCs and geometries, DeepONet suffers from severe overfitting, producing fields with little resemblance to the labels. With varying BCs, it is prone to show the velocity distribution in a steady state while with varying geometries, it tends to behave as identity transformations.

\begin{figure}[!t]
    \centering
    \includegraphics[width=1.0\linewidth]{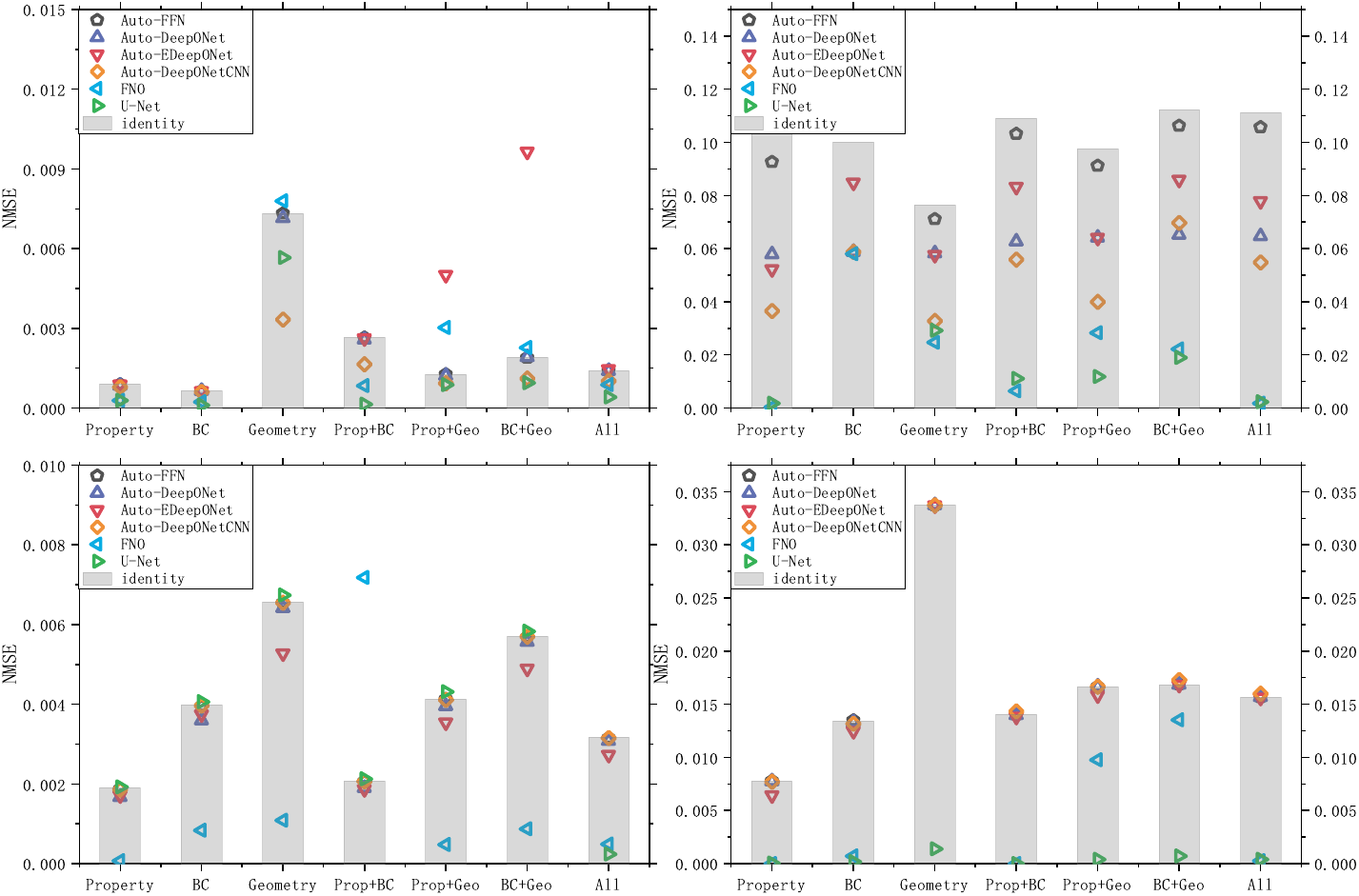}
    \caption{The prediction results of 6 autoregressive baseline models on 7 subsets of data set, with the vertical axis representing the average NMSE of all frames in the test set and the horizontal axis representing the data type.}
    \label{fig:result-auto}
\end{figure}

\begin{figure}[!t]
    \centering
    \includegraphics[width=1.0\linewidth]{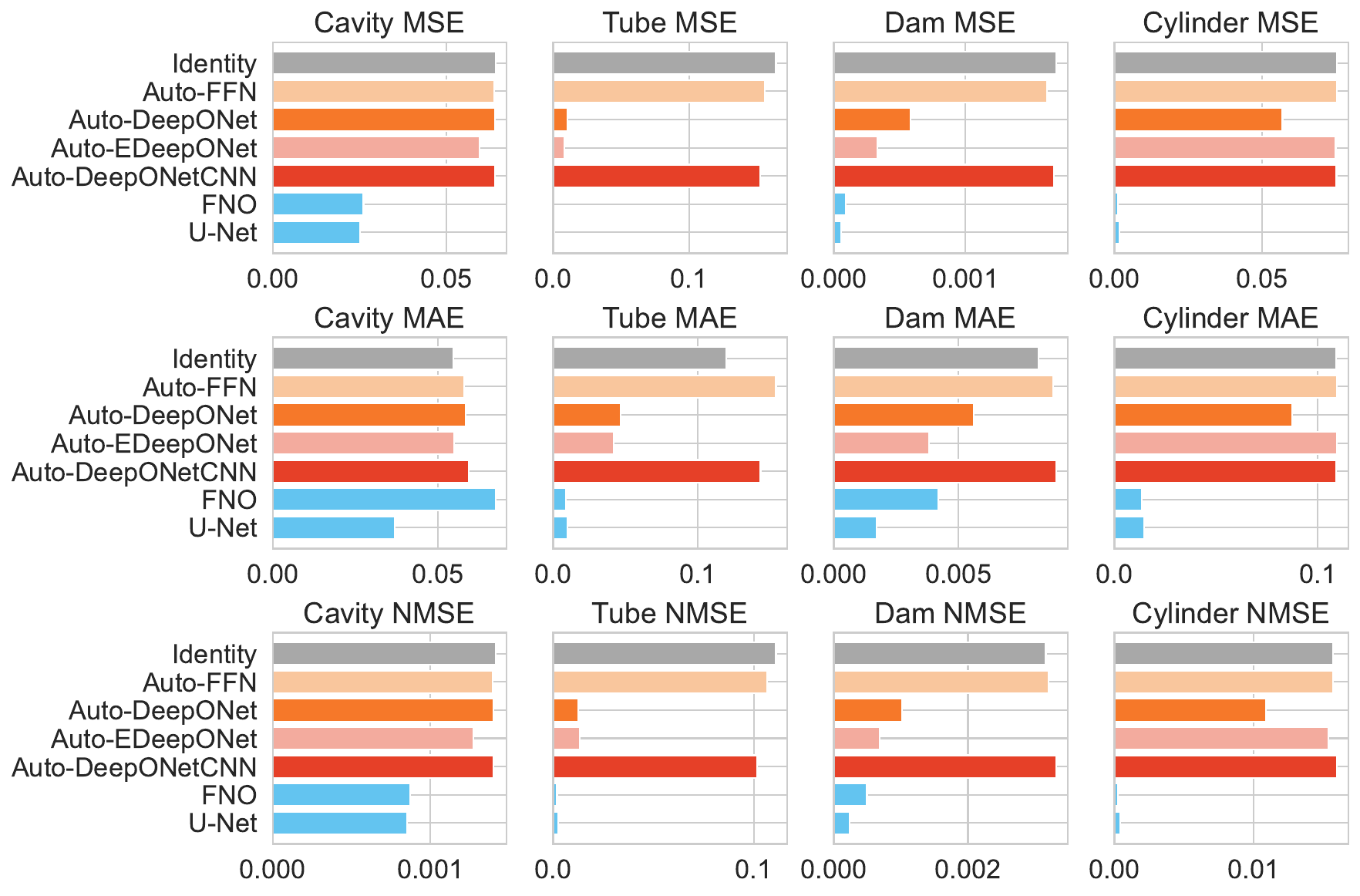}
    \caption{Summary of the performance of autoregressive baseline methods on the four problems (with all cases). FNO's result on the dam flow problem is removed because the error is too large and including it would make the plot less intelligible. The bar chart of the cylinder flow (d) is in logarithmic scale.}
    \label{fig:result-summary}
\end{figure}

Figure~\ref{fig:result-auto} shows the test NMSE of all autoregressive models and the identity transformation on the four flow problems with 7 subsets of data. Figure~\ref{fig:result-summary} shows the test NMSE of the autoregressive models on the four flow problems (with all cases), this serves as a comprehensive summary of the performance of the autoregressive baselines. The complete result of our experiments is listed in given in \ref{sec:detailed-results} which contains the test NMSE, MSE, and MAE of each autoregressive model on each of the seven subsets of the four problems in CFDBench.

In general, Auto-FFN and autoregressive models from the DeepONet family are at best slightly better than the identity transformation, which means they often learn to output the input as their predictions.

In cavity flow and tube flow, U-Net demonstrates superior performance due to its encoding-decoding structure in the spatial domain, which enables it to capture sharp changes in the velocity field more effectively. On the other hand, the MSE of U-Net and FNO is small while the MAE is large. This is because the velocities have generally small absolute values ($u<1$), and the relative error is large when the absolute error is small.

In dam flow prediction, the DeepONet family generally prevails while the non-convergence phenomenon is observed in FNO (FNO's result is excluded from the bar chart because the error is too large). The presence of gravity as a dominant physical force in dam flow suggests that the DeepONet family may be more effective in handling PDEs with source terms. 

Both image-to-image models perform the best in the cylinder flow ($\text{MSE} \sim 10^{-5}$, $\text{MAE} \sim 3 \times 10 ^{-3} $), and in this dataset, FNO is better than U-Net. We conjecture this is because FNO is endowed with an ability to extract the characteristics of the periodic vortex more effectively by learning in the frequency domain. 

For the tube flow problem, U-Net’s predictions have horizontal stripe noises while FNO manifests vertical pattern noise at $t=0$. For the cylinder flow problem, we can see from the prediction that although FNO’s test loss is very low, it produces visible noises. This is because, in FNO, high frequencies are discarded to reduce the computational cost, as a result, it struggles to model flat regions and sharp changes. This also implies that the loss functions we have considered (which are also used in many previous works) may not be good at capturing all the artifacts of various methods. 

\subsection{Multi-Step Prediction}

One important characteristic of traditional numerical methods is that they can extrapolate to any time points through an arbitrary number of forward propagation steps (provided that the iterative process converges). Consequently, it is desirable for data-driven deep learning methods to effectively generalize to time steps beyond those encountered during the training phase and predict the field at any time step. For non-autoregressive models, we can simply query points beyond the temporal range of the training distribution, but for autoregressive models, since predictions depend on the previous predictions, errors may accumulate over multiple forward propagation steps \citep{pangu-weather}.

\begin{figure}[!t]
    \centering
    \includegraphics[width=1.0\linewidth]{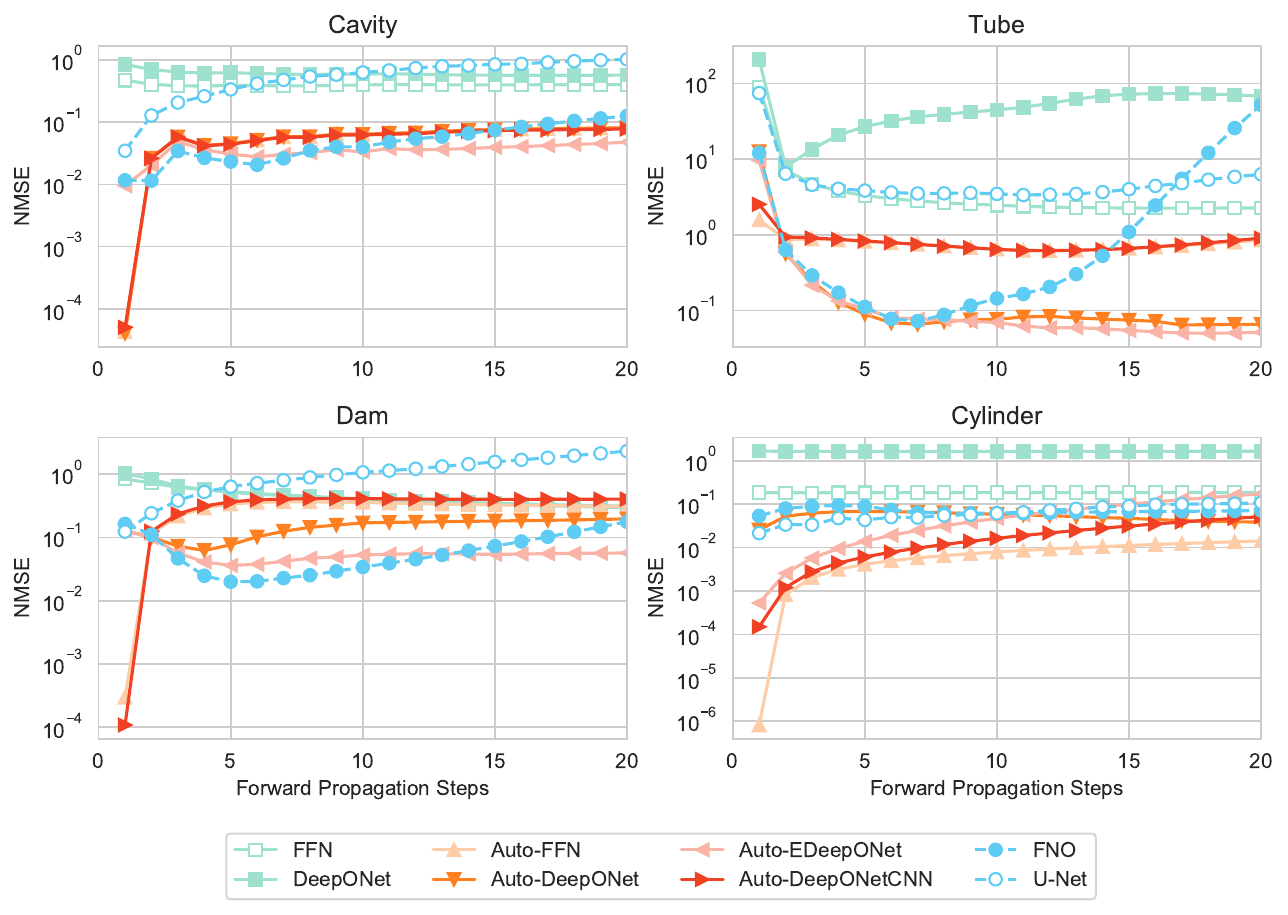} 
    \caption{The error of autoregressive and non-autoregressive baselines as a function of the number of forward propagation steps, given only the operating parameters $\Omega$ and the initial conditions, evaluated on all cases in each task. The $y$-axis is on a logarithmic scale. ResNet is not included because its error is too high and including its line would make the figure less intelligible. Best viewed in color.}
    \label{fig:result-temporal-extrapolation}
\end{figure}

Figure~\ref{fig:result-temporal-extrapolation} illustrates the errors of the baseline models by propagating from the IC (the velocity field at $t=0$) with respect to different time steps. As expected, autoregressive models exhibit severe error accumulation. One illustrative example is observed in the tube flow problem, where FNO's error increases by 1000 times within just 15 forward propagation steps. This adheres to our intuition. Perhaps surprisingly, in some cases such as in the tube flow problem, the errors of autoregressive models decrease over time, which means that the model was able to produce a prediction that is more accurate (with respect to the label) when it is provided a less accurate input function. In other words, some autoregressive models are able to utilize the operating conditions to correct themselves. Another observation is that models with convolutional layers, i.e., U-Net, FNO, and Auto-DeepONetCNN, are more prone to error accumulation than other baselines. One possible explanation is that convolutional layers treat the features at different locations the same way, while fully connected layers have one set of weights dedicated to processing every input point.

Concerning non-autoregressive models, the errors are relatively stable with respect to the time step. However, they are outperformed by autoregressive baseline models most of the time. In the worst case, DeepONet is consistently worse than all other baselines, even after 20 steps of error accumulation.

Mitigating error accumulation is an active research direction and is out of the scope of this paper. One approach, suggested by \citep{pangu-weather}, is to train multiple models with varying step sizes. An alternative strategy involves imposing physical constraints on the model, effectively rendering it ``physics-informed''.


\subsection{Computational Cost}

\begin{table}[htb]
    \centering
    \small
    \caption{Computational cost of the different baseline models on the cavity flow data, PROP subset. training time refers to the time required for training before using the model, inference time refers to the time required for computing one forward propagation (for autoregressive models) or the prediction on one query point (for non-autoregressive models).}

    \begin{tabular}{l|r|rr|rr}
        \toprule
            &
            & \multicolumn{2}{c|}{\textbf{Training}}
            & \multicolumn{2}{c}{\textbf{Inference}} \\
        \textbf{Method} 
            & \textbf{\# Param.} 
            & \textbf{Time (min)}
            & \textbf{Mem. (MB)}
            & \textbf{Time (ms)}
            & \textbf{Mem. (MB) } \\
        \midrule
        \multicolumn{6}{c}{Non-Autoregressive Models}\\
        \midrule
        FFN 
            & 72K
            & 30
            & 443
            & 6.4 
            & 7.2 \\
        DeepONet
            & 143K 
            & 28
            & 355.1
            & 7.2
            & 10.3 \\
        \midrule
        \multicolumn{6}{c}{Autoregressive Models} \\
        \midrule
        Auto-FFN 
            & 1,102K
            & 313
            & 4,127
            & 7.5
            & 135.9\\
        Auto-DeepONet 
            & 552K
            & 15
            & 151
            & 6.0 
            & 5.6 \\
        Auto-EDeepONet 
            & 623K
            & 16
            & 153
            & 6.0
            & 5.9 \\
        Auto-DeepONetCNN 
            & 743K
            & 100
            & 1,367
            & 10.3
            & 38.9 \\
        ResNet
            & 522K
            & 105
            & 1,073
            & 10.5 
            & 5.5\\
        FNO 
            & 1,189K
            & 43
            & 475
            & 9.7
            & 13.6\\
        U-Net 
            & 1,095K
            & 33 
            & 224
            & 11.0 
            & 46.1 \\
        \bottomrule
    \end{tabular}
    \label{tab:result-computational-cost}
\end{table}

For more complex flow problems, traditional numerical methods can be very expensive in terms of computational cost, often requiring days and even months to run simulations. It has been shown that deep learning methods can be multiple orders of magnitude faster than numerical methods \citep{fno,deeponet,pangu-weather}, which is one of the primary advantages of data-driven methods \citep{pinn}. 

Different from traditional numerical methods, deep learning methods also involve a training procedure, which can be very time-consuming, and a set of parameters that can be very memory-consuming. Thus, we need to consider these two aspects in addition to the inference time. We measured the computational cost of each baseline model in terms of time and memory usage during training and inference time. The result is listed in Table~\ref{tab:result-computational-cost}. The models are implemented with PyTorch and executed using GPU. The statistics for training are measured with a batch size of 32, and for inference, we use a batch size of 1. The experiment was conducted with one computed with one i7-12700F CPU and one RTX 3060 GPU.

From the result, we see that different models have very different computational costs, especially during training. Auto-FFN is around 21 times slower than Auto-DeepONet in training, despite having only double the number of parameters and no significant difference in prediction error. This is intuitive because as mentioned in Section~\ref{sec:deeponet}, by reusing the output of the branch net within one mini-batch, DeepONet can significantly improve the training efficiency. Another important observation from this result is that autoregressive models generally have many more parameters compared to non-autoregressive models, but the two kinds of models have comparable training computational costs. This is because autoregressive baselines predict the entire output function with one forward pass, while non-autoregressive baselines predict each data point of the output function one by one.

On the other hand, during inference, the time needed for the models to perform one forward propagation (or one query for non-autoregressive models) is very similar, all within the range of 5 to 10 ms. This is much faster than the numerical method employed for the generation of this dataset, which takes around 1 second for every frame.

\section{Conclusions}

We have introduced CFDBench, a large-scale multi-task benchmark for evaluating the inference-time generalization ability of neural operators in fluid dynamics. CFDBench includes four archetypal flow scenarios: (1)~the lid-driven rectangular cavity flow, (2)~laminar boundary layer flow, (3)~flow resulting from a dam break, and (4)~flow around a cylindrical obstacle. Each problem can be split into three subsets, each containing examples with different (1)~BCs, (2)~fluid physical properties, or (3)~domain geometries. Such variations in operating conditions are designed to examine the ability of data-driven deep learning methods to generalize to unseen conditions without training. 

Secondly, we use the constructed data set to benchmark the ability of mainstream models to predict the distributions of velocity. The models include two non-autoregressive models FFN and DeepONet, and four autoregressive models, namely, the autoregressive version of FFN, DeepONet, EDeepONet, and DeepONetCNN, and three three image-to-image models, namely ResNet, U-Net, and FNO. Before training the model, we introduced each model in detail, compared the differences between the models, and carried out a hyperparameter search for each model to get a suitable hyperparameter for the flow problem. 

By analyzing the single-step and multi-step prediction of baselines on CFDBench, we find that U-Net is the best for the flow problem without source term (gravity), FNO is the best for the phenomenon of periodic eddy currents, and autoregressive DeepONetCNN is the best for the dam flow problem with gravity. The non-autoregressive models with the advantage of grid independence, perform well on the flow problems with relatively small changes in the flow field, such as the cavity flow and the dam flow, but they struggle to converge on the tube flow and cylinder flow problem. In the results of multi-step inference, the fully-connected neural networks are significantly better than the convolutional neural network, non-autoregressive models consistently outperform autoregressive models. The root-mean-square error eventually becomes stable with the extension of the extrapolation time.

All the results of this article show that although these methods perform well on simple problems, they exhibit limited generalization ability on more challenging dimensions, and thus there is still much room for improvement. We are convinced that our dataset provides an important first step towards better designing data-driven neural operators for CFD.

\section*{Data Availability}

Data will be made available upon request.

\bibliographystyle{plain}
\bibliography{refs}

\clearpage
\appendix

\section{Mathematical Notations}

For clarity, we list all commonly used mathematical notations and the corresponding definitions in Table~\ref{tab:math-notations}.

\begin{table}[H]
    \centering
    \caption{Definition of common mathematical notations used in the paper.}
    \begin{tabular}{c|p{66mm}}
        \toprule
        \textbf{Nota.} & \textbf{Definition} \\
        \midrule
        $\nabla$        & The differential operator.\\
        $\mathcal D$    & The domain of the fluid field. \\
        $T$             & The maximum time step of interest. \\
        $\Delta t$      & The time difference between two adjacent time frames. \\
        $\mathbf{u}$    & The velocity of the fluid. \\
        $u_\mathcal B$  & The boundary conditions. \\ 
        $u_\text{sample}$ & The set of query points on the input field function. \\
        $u$             & The x-velocity of the fluid. \\
        $v$             & The y-velocity of the fluid. \\
        $\tau$          & The shearing stress of fluid. \\
        $\rho$          & Density of fluid.\\
        $\mu$           & Viscosity of fluid.\\
        $\Omega$        & The working condition parameters, which is $(u_\mathcal B, \rho, \mu, S)$, where $S$ denotes the shape of the spatial domain, which is different for each problem. \\
        $\Sigma$        & Input function to the PDE solver. \\
        $\theta$        & The parameters of a neural network. \\
        $f_\theta$      & A neural model parameterized by $\theta$. \\
        $\mathcal L$    & The training loss function. \\
        $\mathcal T$    & The training data. \\
        $\mathbf{Y}$ & The label value of training data. \\
        $\hat{\mathbf{Y}}$ & The predicted value of training data. \\     
        $||$            & Concatenation operator \\
        $f_B$           & The branch net in DeepONet. \\
        $f_T$           & The trunk net in DeepONet. \\
        $b$             & The bias term in DeepONet. \\
        \bottomrule
    \end{tabular}
    \label{tab:math-notations}
\end{table}

\section{Detailed Baseline Models}
\label{sec:baseline}

\subsection{Non-Autoregressive Baselines}
\label{sec:non-autoregressive-baselines}

In non-autoregressive modeling, we refer to the operating condition $\Omega$ as the input function.

\subsubsection{FFN}

FFN (feed-forward network) is the simplest form of non-autoregressive modeling. The coordinates of the query location and the input function are simply concatenated into one vector, and fed to a chain of fully connected layers, with a non-linear function after every layer except for the last layer. Thus, the prediction is
\begin{align}
\hat u(x, y, t) = f_\theta(\Omega || (x, y, t)),
\end{align}
where $||$ is the concatenation operator. This model is depicted in Figure \ref{fig:networks-structure}a and it can be regarded as the data-driven version of PINN \citep{pinn}.

\subsubsection{DeepONet}
\label{sec:deeponet}

\citep{deeponet} have shown that by separating the encoding process of the input function and the query location can reduce error. They are encoded by two separate FFNs, the branch net and the trunk net. The outputs are aggregated by dot product to produce the final prediction:

\begin{equation}
    \hat u(x,y,t) = f_B(\Omega) \cdot f_T(x, y, t) + b,
\end{equation}

where $f_B$ and $f_T$ are the branch and trunk net, and $b\in \mathbb R$ is a trainable scalar that acts as the bias term. In other words, DeepONet is a specific case of FFN where each linear layer is cut in half, and each neuron can only see the operating parameters $\Omega$ or only the query coordinates $(x,y,t)$.

Furthermore, to improve the training speed of DeepONet, we can reuse the output of the branch net within each mini-batch. We sample $k=1000$ points in each frame as labels. $f_B(\Omega)$ is computed once, and each of the 1000 points $(x,y,t)$ are dotted with $f_B(\Omega)$ before updating the model weights. Figure~\ref{fig:networks-structure}b illustrates the structure of DeepONet.

\subsection{Autoregressive Baselines}
\label{sec:autoregressive-baselines}

Autoregressive is arguably more similar to traditional numerical solvers, where the model predicts the flow state at the next time step given the previous time step, i.e., $f_\theta: (u (t - \Delta t), \Omega) \mapsto u(t)$. Image-to-image models directly model $f_\theta $, and different image-to-image models differ only in the implementation of $f_\theta$.

\subsubsection{Autoregressive FFN}

The autoregressive FFN is similar to the non-autoregressive version. The input field, operating conditions, and the query location are all concatenated and fed to an FFN, which predicts the current field at the query location:

\begin{align}
   \hat u(x,y,t) = f_\theta \left( u_\text{sample} || \Omega || (x, y) \right ) ,
\end{align}

where $u_\text{sample}$ refers to a list of field values sampled from $u$. This can be seen as a completely data-driven version of PINN \citep{pinn}. Figure~\ref{fig:networks-structure}a depicts the structure of Auto-FFN.

\subsubsection{Autoregressive DeepONet}

We also consider modifying DeepONet to allow it to generate the solution autoregressively, and we name this model \textbf{Auto-DeepONet}. The structure is shown in Figure~\ref{fig:networks-structure}c. The input to the branch net (i.e., the input function) is $(u(t-\Delta t),\Omega)$ where $u(t-\Delta t)$ is the last predicted velocity field and $\Omega$ is the operating condition parameters. The input to the trunk net is the spatial coordinates $(x,y)$ of the query location, while the target output of the model is the value of the velocity field in the next time frame at $(x,y)$, i.e., $u(x,y,t)$. The model is formulated as follows.

\begin{equation*}
    \hat u(x, y, t) = f_B \left(u_\text{sample}(t - \Delta t) || \Omega \right) \cdot f_T(x, y) + b
\end{equation*}

\subsubsection{Autoregressive EDeepONet}

EDeepONet (Enhanced DeepONet)\citep{edeeponet} extends DeepONet's architecture to consider multiple input functions. EDeepONet has one branch net for encoding each input function independently, and the branch outputs are aggregated by element-wise product. Since in autoregression, the DeepONet conditions on two inputs, $u(t - \Delta t)$ and $\Omega$, we also evaluate the autoregressive version of, \textbf{Auto-EDeepONet}. The prediction is modeled as follows.

\begin{equation}
\hat u(x, y, t) = 
\left[ 
    f_{B1} \left(
        u_\text{sample}(t - \Delta t)
    \right) \odot f_{B2} \left( \Omega \right)
\right ] \cdot f_T(x, y) + b
\end{equation}

where $\odot$ denotes the element-wise product.

In other words, EDeepONet is a specific case of DeepONet, where the branch net is split into two parts, each responsible for one input functions) and the neural links between each piece are removed (or deactivated by setting them to zero). This structure is illustrated in Figure~\ref{fig:networks-structure}d.

We do not evaluate the non-autoregressive version of EDeepONet because our preliminary experiments show that splitting $\Omega$ has no significant impact on the ability of the neural network. However, in autoregression, the input includes $u_\text{sample}(t-\Delta t)$, which is much larger than $\Omega$, and simply concatenating the two vectors may cause the neural to fail to learn dependence on $\Omega$.

\subsubsection{Autoregressive DeepONetCNN}

We also experimented with CNN as the feature extractor for the input field called \textbf{Auto-DeepONetCNN}. This is almost the same as Auto-DeepONet, but the $f_B$ is implemented with a CNN, because CNN may be better at extracting features from a lattice of a field. Since CNN requires a cuboid input, the input to the branch net needs to be $u(t - \Delta t)$ instead of $u_\text{sample}(t-\Delta t)$. Similar to ResNet, U-Net, and FNO, $\Omega$ is appended to $u(t-\Delta t)$ as additional channels. The formulation is as follows.

\begin{equation}
    \hat u(t) = \text{CNN}(u(t- \Delta t), \Omega) \odot f_T(x, t) + b 
\end{equation}

\subsubsection{ResNet}

A residual neural network (ResNet) is a CNN with residual connections proposed by \citep{resnet}, and it has shown excellent performance on many computer vision tasks. Residual connectivity can effectively alleviate the degradation problem of the neural network when the depth increases, thus enhancing the learning ability of the model.\footnote{Interestingly, a ResNet block adds the input to the output of a CNN block, which is $x+f(x)$ and is similar to many iterative numerical methods.} The model can be formalized as follows.\footnote{$f\circ g$ denotes the composite of $f$ and $g$.}

\begin{align}
    &\hat u (t) = \text{ResNetBlock}_l \circ \dots \circ \text{ResNetBlock}_1(x) \\
    &x = f_{in}(u(t-\Delta t), \Omega) \\
    &\text{ResNetBlock}_i(x) = x + \text{CNN}_i(x)   \quad \quad i = 1, \dots, l
\end{align}

where $\text{CNN}(\cdot)$ is a CNN network.

ResNet has many possible ways to put the ResNet blocks together, and this paper uses a string of residual blocks of the same size.

\subsubsection{U-Net}

U-Net \citep{unet} is a CNN with an encoder-decoder structure, which performs very well in numerous image segmentation and image-to-image translation tasks. The encoder realizes feature extraction and parameter reduction through down-sampling methods such as convolution (with larger striding) and pooling, and the decoder uses the feature encodings to produce an image through up-sampling and channel splicing, so as to achieve the purpose of image generation or segmentation. Compared to ResNet, the down-sampling of U-Net can reduce the number of parameters. Up-sampling can improve the globality of the convolution kernel because after up-sampling, a signal affects a larger region than without up-sampling. The structure of the U-Net used in this paper is illustrated in Figure~\ref{fig:networks-structure}f.

\begin{align}
   & \hat u(t)  = \text{UNet}(u(t-\Delta t), \Omega) && \hat y    = f_{out} (y_0) \\
  &  x_0     = f_\text{in} (u(t-\Delta t), \Omega)  && y_0       = \text{UpConv}_1 (y_{1}, x_0)  \\
   & x_1     = \text{DownConv}_1(x_{0})             && y_1       = \text{UpConv}_2 (y_{2}, x_1) \\
            &\vdots                                 && \vdots        \\
   & x_{l}     = \text{DownConv}_{l-1}(x_{l-1})     && y_l       = \text{UpConv}_l (y_{l+1}, x_{l}) \\
  &  x_{l+1}     = \text{DownConv}_l(x_{l})         && y_{l+1}   = x_{l+1}
\end{align}

where

\begin{align}
    &\text{DownConv}(x_i)) = \text{Conv}(\text{DownSample}(x_{i-1}))  \\
    &\text{UpConv}(y_i, x_j) = \text{Conv}(\text{UpSample}[y_{i+1}] || x_i) & i = 1, \cdots, l,
\end{align}

$\text{Conv}_i(\cdot)$ denotes a CNN, $\text{UpSample} (\cdot)$ and $\text{DownSample} (\cdot)$ denotes the up-sampling and down-sampling functions, $l$ denotes the number of U-Net blocks, and $f_{in}(\cdot)$ and $f_{out}(\cdot)$ denotes two trainable mappings. This architecture is shown in Figure~\ref{fig:networks-structure}f.

\subsubsection{FNO}

Fourier neural operator (FNO)\citep{fno} is a neural network that parameterizes the convolution kernel in Fourier space. It can learn the mapping of high-dimensional space and especially performs well in the problem of turbulent pulsation. The Fourier neural operator first raises the input function to a high-dimensional space through a shallow fully connected network and then approaches the target transform through the Fourier layer containing the Fourier transform and the inverse transform. The FNO has better globality than an ordinary CNN because any signal in the Fourier space affects the output on the entire spatial domain. Figure~\ref{fig:networks-structure}e shows the structure of FNO, and it can be formalized as follows.

\begin{align}
    \hat u(t) &= Q(h_{l+1}) \\
    h_{i+1} = \text{FourierBlock}_i(h_{i}) 
        &= \mathcal F^{-1} [ R_i ( \mathcal F [h_{i}] ) ] + W_i(h_i) + h_i & i= 1, \dots, l \\
    h_1 &= P(u(t-\Delta t), \Omega)
\end{align}

where $\mathcal F(\cdot)$ denotes the Fourier transform, $P(\cdot)$, $Q(\cdot)$, $W_i(\cdot)$ are ordinary convolutional layers, and $R_i(\cdot)$ is a $1 \times 1$ convolutional layer.

It is worth mentioning that, in the original paper of FNO \citep{fno}, the input includes multiple time steps before the current time step, which provides additional information about the flow's state and may make inference easier. However, this limits the usability of the method. Therefore, in this work, we only consider the scenario where the input contains no more than one frame.

\section{Detailed Results}
\label{sec:detailed-results}

\subsection{None-autoregressive Baselines}
Here, we list the detailed results of the autoregressive baselines on the four problems in CFDBench, shown in Table \ref{tab:non-auto-results}

\begin{table*}[h]
    \centering
    \footnotesize
    \caption{Main test results of non-autoregressive methods (FFN and DeepONet) on each subset of each problem.}
            \begin{tabular}{c|l|cc|cc|cc}
                \toprule
                \multirow{2}{*}{Problem}
                    & \multirow{2}{*}{Subset}
                    & \multicolumn{2}{c|}{NMSE}
                    & \multicolumn{2}{c|}{MSE}
                    & \multicolumn{2}{c}{MAE}
                \\
                    &
                    & FFN & DeepONet
                    & FFN & DeepONet
                    & FFN & DeepONet
                \\
                \midrule
                \multirow{7}{*}{Cavity}
                
                & \textbf{(1) PROP}
                    & \textbf{0.0099592} & 0.0291865
                    & \textbf{0.0473762} & 0.1693782
                    & \textbf{0.1111506} & 0.5680576\\
                & \textbf{(2) BC}
                    & \textbf{0.0023445} & 0.1110036
                    & \textbf{0.2437581} & 13.2787848
                    & \textbf{0.2882956} & 1.7553163\\
                & \textbf{(3) Geo}
                    & 0.6239881 & \textbf{0.5806319}
                    & \textbf{1.8697622} & 1.9919338
                    & \textbf{0.9227801} & 0.9277460\\
                & \textbf{(4) PROP + BC}
                    & \textbf{0.0084082} & 0.0799971
                    & \textbf{0.2003079} & 3.3371139
                    & \textbf{0.2580179} & 0.8954092\\
                & \textbf{(5) PROP + GEO}
                    & 0.1242569 & \textbf{0.0892609}
                    & 0.3536154 & \textbf{0.3205143}
                    & \textbf{0.2056559} & 0.2610952\\
                & \textbf{(6) BC + GEO}
                    & \textbf{0.1161350} & 0.2098982
                    & \textbf{1.1426576} & 10.0195319
                    & \textbf{0.5412614} & 1.5829833\\
                & \textbf{(7) All}
                    & \textbf{0.0221644} & 0.0646872
                    & \textbf{0.7857684} & 4.0079125
                    & \textbf{0.4356092} & 1.0548950\\

                \midrule
                \multirow{7}{*}{Tube}
                & \textbf{(1) PROP}
                    & \textbf{0.0004197} & 0.0039994
                    & \textbf{0.0002522} & 0.0026946
                    & \textbf{0.0078708} & 0.0315951\\
                & \textbf{(2) BC}
                    & \textbf{19.2247428} & 25.8505255
                    & 1.4869019 & \textbf{1.3689488}
                    & 0.7780905 & \textbf{0.7297134}\\
                & \textbf{(3) GEO}
                    & \textbf{0.1674582} & 0.1861845
                    & 0.1735853 & \textbf{0.1708268}
                    & \textbf{0.2698841} & 0.3242756\\
                & \textbf{(4) PROP + BC}
                    & \textbf{3.7321264} & 5.8203221
                    & 0.5254855 & \textbf{0.4897547}
                    & 0.3910940 & \textbf{0.3403379}\\
                & \textbf{(5) PROP + GEO}
                    & 0.6573232 & \textbf{0.5961287}
                    & \textbf{0.1125305} & 0.1187899
                    & \textbf{0.1520187} & 0.1924667\\
                & \textbf{(6) BC + GEO}
                    & 0.6412595 & \textbf{0.5119403}
                    & \textbf{1.7430317} & 2.1646790
                    & \textbf{0.8040325} & 0.8678228\\
                & \textbf{(7) All}
                    & 3.0935680 & \textbf{0.3437079}
                    & 0.5307588 & \textbf{0.4553377}
                    & 0.3303886 & \textbf{0.3416610}\\

                \midrule
                \multirow{7}{*}{Dam} 
                & \textbf{(1) PROP}
                    & \textbf{0.0004000} & 0.0205820
                    & \textbf{0.0002025} & 0.0104605
                    & \textbf{0.0080647} & 0.0602617\\
                & \textbf{(2) BC}
                    & 0.3882083 & \textbf{0.0145171}
                    & 0.1656223 & \textbf{0.0098575}
                    & 0.2962269 & \textbf{0.0512015}\\
                & \textbf{(3) GEO}
                    & \textbf{0.0408200} & 0.0438950
                    & \textbf{0.0101389} & 0.0109773
                    & \textbf{0.0449910} & 0.0517545\\
                & \textbf{(4) PROP + BC}
                    & 0.3830672 & \textbf{0.0048370}
                    & 0.0522189 & \textbf{0.0019914}
                    & 0.1178206 & \textbf{0.0223580}\\
                & \textbf{(5) PROP + GEO}
                    & \textbf{0.0190430} & 0.0567282
                    & \textbf{0.0060118} & 0.0231003
                    & \textbf{0.0319056} & 0.0772180\\
                & \textbf{(6) BC + GEO}
                    & \textbf{0.0982004} & 0.3650784
                    & \textbf{0.0431119} & 0.1924977
                    & \textbf{0.1442159} & 0.3337956\\
                & \textbf{(7) All}
                    & 0.1694195 & \textbf{0.0686736}
                    & 0.0705092 & \textbf{0.0270029}
                    & 0.1362603 & \textbf{0.1007961}\\
                
                \midrule
                \multirow{7}{*}{Cylinder}
                
                & \textbf{(1) PROP}
                    & \textbf{0.0007879} & 0.0021776
                    & \textbf{0.0008786} & 0.0024212
                    & \textbf{0.0141193} & 0.0254937\\
                & \textbf{(2) BC}
                    & \textbf{0.0108285} & 9.7361195
                    & \textbf{0.0682347} & 4.0023353
                    & \textbf{0.1358656} & 1.5573399\\
                & \textbf{(3) GEO}
                    & \textbf{0.1405526} & 108.5875535
                    & \textbf{0.1648840} & 119.6764528
                    & \textbf{0.2541922} & 5.7167007\\
                & \textbf{(4) PROP + BC}
                    & 0.8656293 & \textbf{0.2141155}
                    & 0.9652876 & \textbf{0.4134728}
                    & \textbf{0.2702630} & 0.4543390\\
                & \textbf{(5) PROP + GEO}
                    & \textbf{0.0249946} & 0.1252280
                    & \textbf{0.0290181} & 0.1412260
                    & \textbf{0.0731633} & 0.2759877\\
                & \textbf{(6) BC + GEO}
                    & \textbf{0.0560368} & 0.0570367
                    & \textbf{0.0899771} & 0.0966049
                    & 0.1741357 & \textbf{0.1707578}\\
                & \textbf{(7) All}
                    & \textbf{0.0281058} & 2.3627123
                    & \textbf{0.0472888} & 3.0325988
                    & \textbf{0.1155052} & 1.2104118\\

                \bottomrule
            \end{tabular}
    \label{tab:non-auto-results}
\end{table*}

\subsection{Autoregressive Baselines}

Here, we list the detailed results of the autoregressive baselines on the four problems in CFDBench, shown in Table \ref{tab:auto-results-cavity}, \ref{tab:auto-results-tube}, \ref{tab:auto-results-dam}, and \ref{tab:auto-results-cylinder}.

\begin{table}[htbp]
    \footnotesize
    \centering
        \caption{Detailed results of autoregressive baseline models on the Cavity Flow problem on one forward propagation.}
        \begin{tabular}[c]{l|ccccccc}
            \multicolumn{8}{l}{Problem 1: Cavity Flow} \\
            \toprule
            \textbf{Method}
                & \textbf{(1) PROP} 
                & \textbf{(2) BC} 
                & \textbf{(3) GEO} 
                & \textbf{(4) P+B }
                & \textbf{(5) P+G} 
                & \textbf{(6) B+G} 
                & \textbf{(7) All} \\
            \midrule
            \multicolumn{8}{c}{Test NMSE} \\
            \midrule
            Identity 
                & 0.0008949 
                & 0.0006532 
                & 0.0073354 
                & 0.0026440 
                & 0.0012782 
                & 0.0019072 
                & 0.0014130\\
            Auto-FFN 
                & 0.0008947 
                & 0.0006536 
                & 0.0073358 
                & 0.0026441 
                & 0.0012785 
                & 0.0019086 
                & 0.0014138\\
            Auto-DeepONet 
                & 0.0008465 
                & 0.0006478 
                & 0.0071480 
                & 0.0025767 
                & 0.0012198 
                & 0.0019119 
                & 0.0013954\\
            Auto-EDeepONet 
                & 0.0008953 
                & 0.0006539 
                & 0.0239769 
                & 0.0026405 
                & 0.0050122 
                & 0.0096511 
                & 0.0014756\\
            Auto-DeepONetCNN 
                & 0.0007973 & 0.0006152 & \textbf{0.0033303} & 0.0016539 & 0.0009240 & 0.0011091 & 0.0010203\\
            FNO 
                & 0.0002864 
                & 0.0002247 
                & 0.0078046
                & 0.0008500
                & 0.0030215 
                & 0.0022841
                & 0.0008715\\
            U-Net 
                & \textbf{0.0002815} 
                & \textbf{0.0001159} 
                & 0.0056645 
                & \textbf{0.0001383} 
                & \textbf{0.0008825} 
                & \textbf{0.0009481} 
                & \textbf{0.0004166}\\

            \midrule
            \multicolumn{8}{c}{Test MSE} \\
            \midrule
            Identity 
                & 0.0044942 
                & 0.0546373 
                & 0.0180946 
                & 0.0990002 
                & 0.0045866 
                & 0.0714307 
                & 0.0641640\\
            Auto-FFN 
                & 0.0044936 
                & 0.0546386 
                & 0.0180943 
                & 0.0990127 
                & 0.0045873 
                & 0.0714420 
                & 0.0641692\\
            Auto-DeepONet 
                & 0.0042624 
                & 0.0536015 
                & 0.0179865 
                & 0.0976493 
                & 0.0044118 
                & 0.0710621 
                & 0.0638475\\
            Auto-EDeepONet 
                & 0.0044994 & 0.0547824 & 0.0669539 & 0.0989851 & 0.0156476 & 0.0950095 & 0.0644919\\
            Auto-DeepONetCNN 
                & 0.0043076 & 0.0531823 & 0.0125515 & 0.0920075 & 0.0043748 & 0.0709266 & 0.0632696\\
            FNO 
                & \textbf{0.0012038} & 0.0109657 & 0.0196869 & 0.0202371 & 0.0080227 & 0.0361190 & 0.0261403\\
            U-Net 
                & 0.0044942 
                & \textbf{0.0083046} 
                & \textbf{0.0118261} 
                & \textbf{0.0064567} 
                & \textbf{0.0017226} 
                & \textbf{0.0210355} 
                & \textbf{0.0158059}\\
            \midrule
            \multicolumn{8}{c}{Test MAE} \\
            \midrule
            Identity 
                & 0.0181955 & 0.0506039 & 0.0297850 & 0.0850075 & 0.0181359 & 0.0564395 & 0.0546747\\
            Auto-FFN 
                & 0.0182054 & 0.0507490 & 0.0298784 & 0.0854521 & 0.0183282 & 0.0570768 & 0.0552039\\
            Auto-DeepONet 
                & 0.0192833 & 0.0540527 & 0.0327312 & 0.0814217 & 0.0198544 & 0.0663280 & 0.0566274\\
            Auto-EDeepONet 
                & 0.0200762 & 0.0591863 & 0.1748586 & 0.0869751 & 0.0522350 & 0.0814745 & 0.0600075\\
            Auto-DeepONetCNN 
                & 0.0210971 & 0.0542496 & \textbf{0.0222548} & 0.0792672 & 0.0201459 & 0.0571482 & 0.0584715\\
            FNO 
                & 0.0150832 & 0.0485077 & 0.0553802 & 0.0633005 & 0.0316786 & 0.0735833 & 0.0673980\\
            U-Net 
                & \textbf{0.0103330} 
                & \textbf{0.0001159} 
                & 0.0328206 
                & \textbf{0.0422648} 
                & \textbf{0.0155698} 
                & \textbf{0.0325585} 
                & \textbf{0.0319145}\\
            \bottomrule
        \end{tabular}
        \label{tab:auto-results-cavity}
\end{table}

\begin{table}[htbp]
    \footnotesize
    \centering
        \caption{Detailed results of autoregressive baselines models on the Tube Flow problem on one forward propagation.}
        \begin{tabular}[c]{l|ccccccc}
            \multicolumn{8}{l}{Problem 2: Tube Flow}\\
    
            \toprule
            \textbf{Method}
                & \textbf{(1) PROP} 
                & \textbf{(2) BC} 
                & \textbf{(3) GEO} 
                & \textbf{(4) P+B }
                & \textbf{(5) P+G} 
                & \textbf{(6) B+G} 
                & \textbf{(7) All} \\
            \midrule
            \multicolumn{8}{c}{Test NMSE} \\
            \midrule
            Identity 
                & 0.1081580 & 0.1001696 & 0.0763603 & 0.1089607 & 0.0976491 & 0.1122125 & 0.1111430\\
            Auto-FFN 
                & 0.0926980 
                & 0.1363334 
                & 0.0712057 
                & 0.1032522 
                & 0.0912989 
                & 0.1062881 
                & 0.1056823\\
            Auto-DeepONet 
                & 0.0579279 
                & 0.0587133 
                & 0.0582056 
                & 0.0627424 
                & 0.0642253 
                & 0.0652362 
                & 0.0647747\\
            Auto-EDeepONet 
                & 0.0523948 
                & 0.0849620 
                & 0.0577905 
                & 0.0833847 
                & 0.0641345 
                & 0.0860665 
                & 0.0778912\\
            Auto-DeepONetCNN 
                & 0.0366433 
                & 0.0588061 
                & 0.0327204 
                & 0.0559905 
                & 0.0399490 
                & 0.0696541 
                & 0.0548516\\
            FNO 
                & \textbf{0.0004093} & \textbf{0.0580814} & \textbf{0.0246608} & \textbf{0.0064176} & 0.0283027 & 0.0222346 & \textbf{0.0018324} \\
            U-Net 
                & 0.0018705 
                & 5.0228938 
                & 0.0291472 
                & 0.0111089 
                & \textbf{0.0118453} 
                & \textbf{0.0190382} 
                & 0.0031894\\
            \midrule
            \multicolumn{8}{c}{Test MSE} \\
            \midrule
            Identity 
                & 0.0317068 & 0.3432079 & 0.0298840 & 0.1495200 & 0.0287833 & 0.3478090 & 0.1642216\\
            Auto-FFN 
                & 0.0279299 & 0.3017316 & 0.0280562 & 0.1298233 & 0.0259540 & 0.3374500 & 0.1554814\\
            Auto-DeepONet 
                & 0.0169327 & 0.1229224 & 0.0223923 & 0.0635457 & 0.0189828 & 0.1395774 & 0.0723492\\
            Auto-EDeepONet 
                & 0.0165697 & 0.2007642 & 0.0209080 & 0.0929376 & 0.0175065 & 0.2476731 & 0.0973665\\
            Auto-DeepONetCNN 
                & 0.0268636 & 0.2177070 & 0.0266211 & 0.1133375 & 0.0248359 & 0.2599603 & 0.1031608\\
            FNO 
                & \textbf{0.0000534} 
                & \textbf{0.0014329} 
                & \textbf{0.0052453} 
                & \textbf{0.0003632} 
                & \textbf{0.0022952} 
                & \textbf{0.0063454} 
                & \textbf{0.0002510} \\
            U-Net 
                & 0.0007242 
                & 0.3389257 
                & 0.0132874 
                & 0.0072537 
                & 0.0026152 
                & 0.0142700 
                & 0.0012903\\
        
            \midrule
            \multicolumn{8}{c}{Test MAE} \\
            \midrule
            Identity 
                & 0.0762089 & 0.1662700 & 0.0577343 & 0.1201217 & 0.0673662 & 0.1670072 & 0.1198109\\
            Auto-FFN 
                & 0.1157967 & 0.2040521 & 0.0699715 & 0.1559224 & 0.1017113 & 0.2031115 & 0.1568468\\
            Auto-DeepONet 
                & 0.0764206 & 0.1481193 & 0.0634907 & 0.1215821 & 0.0713417 & 0.1534080 & 0.1195835\\
            Auto-EDeepONet 
                & 0.0754687 & 0.1766330 & 0.0685437 & 0.1341049 & 0.0691925 & 0.1905831 & 0.1323233\\
            Auto-DeepONetCNN 
                & 0.1044903 & 0.2373263 & 0.0796258 & 0.1626967 & 0.0888877 & 0.2177044 & 0.1347066\\
            FNO 
                & \textbf{0.0039848} 
                & \textbf{0.0220363} 
                & \textbf{0.0284487} 
                & \textbf{0.0111299} 
                & \textbf{0.0161863} 
                & \textbf{0.0302296} 
                & \textbf{0.0087568} \\
            U-Net 
                & 0.0139124 
                & 0.4283762 
                & 0.0431357 
                & 0.0349704 
                & 0.0169491 
                & 0.0526696 
                & 0.0181517\\
            \bottomrule
        \end{tabular}
        \label{tab:auto-results-tube}
\end{table}

\begin{table}[htbp]
    \footnotesize
    \centering
        \caption{Detailed results of autoregressive baselines models on the Dam Flow problem on one forward propagation.}
        \begin{tabular}[c]{l|ccccccc}
            \multicolumn{8}{l}{Problem 3: Dam Flow}\\

            \toprule
            \textbf{Method}
                & \textbf{(1) PROP} 
                & \textbf{(2) BC} 
                & \textbf{(3) GEO} 
                & \textbf{(4) P+B }
                & \textbf{(5) P+G} 
                & \textbf{(6) B+G} 
                & \textbf{(7) All} \\
            \midrule
            \multicolumn{8}{c}{Test NMSE} \\
            \midrule
            Identity 
                & 0.0019018 & 0.0039803 & 0.0065650 & 0.0020840 & 0.0041362 & 0.0056979 & 0.0031620\\
            Auto-FFN 
                & 0.0018699 & 0.0039597 & 0.0065501 & 0.0020618 & 0.0041344 & 0.0056924 & 0.0031543\\
            Auto-DeepONet 
                & 0.0016760 & 0.0036014 & 0.0064154 & 0.0019039 & 0.0039516 & 0.0055705 & 0.0030798\\
            Auto-EDeepONet 
                & 0.0017231 & 0.0037461 & 0.0052787 & 0.0018735 & 0.0035536 & 0.0048973 & 0.0027361\\
            Auto-DeepONetCNN 
                & 0.0018616 & 0.0039617 & 0.0065470 & 0.0020625 & 0.0041093 & 0.0057000 & 0.0031518\\
            FNO 
                & \textbf{0.0000729} & \textbf{0.0008408} & \textbf{0.0010921} & \textbf{0.0071788} & \textbf{0.0004772} & \textbf{0.0008742} & \textbf{0.0004927}\\
            U-Net 
                & 0.0019239 & 0.0040659 & 0.0067332 & 0.0021257 & 0.0043181 & 0.0058265 & 0.0032407\\
            
            \midrule
            \multicolumn{8}{c}{Test MSE} \\
            \midrule
            Identity 
                & 0.0011082 & 0.0048598 & 0.0015390 & 0.0016745 & 0.0013234 & 0.0035683 & 0.0016937\\
            Auto-FFN 
                & 0.0010898 & 0.0048156 & 0.0015359 & 0.0016551 & 0.0013186 & 0.0035535 & 0.0016800\\
            Auto-DeepONet 
                & 0.0009789 & 0.0044238 & 0.0015048 & 0.0015351 & 0.0012433 & 0.0034572 & 0.0016390\\
            Auto-EDeepONet 
                & 0.0010059 & 0.0045758 & 0.0012384 & 0.0014951 & 0.0011442 & 0.0031783 & 0.0014936\\
            Auto-DeepONetCNN 
                & 0.0010854 & 0.0048134 & 0.0015349 & 0.0016510 & 0.0013099 & 0.0035586 & 0.0016810\\
            FNO 
                & \textbf{0.0000189} & \textbf{0.0003273} & \textbf{0.0001341} & \textbf{0.0000758} & \textbf{0.0000681} & \textbf{0.0002176} & \textbf{0.0000875} \\
            U-Net 
                & 0.0011090 & 0.0048536 & 0.0015478 & 0.0016708 & 0.0013723 & 0.0035743 & 0.0017016\\
            
            \midrule
            \multicolumn{8}{c}{Test MAE} \\
            \midrule
            Identity 
                & 0.0083432 & 0.0137022 & \textbf{0.0058706} & 0.0087752 & 0.0072361 & 0.0105381 & 0.0082023\\
            Auto-FFN 
                & 0.0075868 & 0.0135726 & 0.0070825 & 0.0079549 & 0.0076244 & 0.0105358 & 0.0082106\\
            Auto-DeepONet 
                & 0.0064191 
                & 0.0127207 
                & 0.0076071 
                & 0.0072435 
                & 0.0065346 & 0.0098888 & 0.0072718\\
            Auto-EDeepONet 
                & 0.0069405 & 0.0124701 & 0.0060389 & 0.0080457 & 0.0062325 & 0.0100111 & 0.0070318\\
            Auto-DeepONetCNN 
                & 0.0073756 & 0.0134733 & 0.0063454 & 0.0082490 & 0.0069973 & 0.0108240 & 0.0080630\\
            FNO 
                & \textbf{0.0022640} 
                & \textbf{0.0080930} 
                & \textbf{0.0047965} 
                & \textbf{0.0039099} 
                & \textbf{0.0036704} 
                & \textbf{0.0061927} 
                & \textbf{0.0042026} \\
            U-Net 
                & 0.0088548 
                & 0.0146514 
                & 0.0072130 
                & 0.0094619 
                & 0.0096587 
                & 0.0111258 
                & 0.0092133\\
            \bottomrule
        \end{tabular}
        \label{tab:auto-results-dam}
\end{table}

\begin{table}[htbp]
    \footnotesize
    \centering
        \caption{Detailed results of autoregressive baseline models on the Cylinder Flow problem on one forward propagation.}
        \begin{tabular}[c]{l|ccccccc}
        \multicolumn{8}{l}{Problem 4: Cylinder Flow}\\
        
        \toprule
        \textbf{Method}
            & \textbf{(1) PROP} 
            & \textbf{(2) BC} 
            & \textbf{(3) GEO} 
            & \textbf{(4) P+B }
            & \textbf{(5) P+G} 
            & \textbf{(6) B+G} 
            & \textbf{(7) All} \\
        \midrule
        \multicolumn{8}{c}{Test NMSE} \\
        \midrule
        Identity 
            & 0.0077999 & 0.0134142 & 0.0337257 & 0.0140363 & 0.0166764 & 0.0168646 & 0.0156948\\
        Auto-FFN 
            & 0.0077977 & 0.0134795 & 0.0337259 & 0.0140332 & 0.0166751 & 0.0168684 & 0.0156955\\
        Auto-DeepONet 
            & 0.0077462 & 0.0133213 & 0.0337149 & 0.0139771 & 0.0166488 & 0.0168484 & 0.0156741\\
        Auto-EDeepONet 
            & 0.0064475 & 0.0125065 & 0.0337160 & 0.0138407 & 0.0158704 & 0.0168613 & 0.0156335\\
        Auto-DeepONetCNN 
            & 0.0077035 & 0.0131733 & 0.0337303 & 0.0143524 & 0.0166459 & 0.0172999 & 0.0160131\\
        FNO 
            & \textbf{0.0000073} & 0.0007250 & 0.0405524 & 0.0015898 & 0.0097638 & 0.0135264 & 0.0002535\\
        U-Net 
            & 0.0000482 & \textbf{0.0002006} & \textbf{0.0014008} & \textbf{0.0000140} & \textbf{0.0004029} & \textbf{0.0007282} & \textbf{0.0000216}\\
    
        \midrule
        \multicolumn{8}{c}{Test MSE} \\
        \midrule
        Identity 
            & 0.0085652 & 0.1510596 & 0.0391798 & 0.0495286 & 0.0189442 & 0.0988010 & 0.0754044\\
        Auto-FFN 
            & 0.0085629 & 0.1510848 & 0.0391801 & 0.0495232 & 0.0189429 & 0.0988837 & 0.0754104\\
        Auto-DeepONet 
            & 0.0085082 & 0.1492194 & 0.0391674 & 0.0493322 & 0.0189148 & 0.0981404 & 0.0753445\\
        Auto-EDeepONet 
            & 0.0071054 & 0.1370049 & 0.0391690 & 0.0480996 & 0.0180690 & 0.0987495 & 0.0743118\\
        Auto-DeepONetCNN 
            & 0.0084855 & 0.1467647 & 0.0392012 & 0.0493385 & 0.0189119 & 0.1009318 & 0.0751691\\
        FNO 
            & \textbf{0.0000080} & 0.0034436 & 0.0464464 & 0.0056817 & 0.0111179 & 0.0172855 & 0.0011461 \\
        U-Net 
            & 0.0000517 & \textbf{0.0019469} & \textbf{0.0016252} & \textbf{0.0000391} & \textbf{0.0004649} & \textbf{0.0016478} & \textbf{0.0000549}\\
    
        \midrule
        \multicolumn{8}{c}{Test MAE} \\
        \midrule
        Identity 
            & 0.0429706 & 0.1702698 & 0.1164363 & 0.0904745 & 0.0689821 & 0.1223419 & 0.1090931\\
        Auto-FFN 
            & 0.0435852 & 0.1716940 & 0.1165288 & 0.0909219 & 0.0692674 & 0.1228507 & 0.1091991\\
        Auto-DeepONet 
            & 0.0434741 & 0.1728876 & 0.1167489 & 0.0912146 & 0.0691589 & 0.1242880 & 0.1095553\\
        Auto-EDeepONet 
            & 0.0430859 & 0.1663353 & 0.1165038 & 0.0903266 & 0.0681725 & 0.1224469 & 0.1095621\\
        Auto-DeepONetCNN 
            & 0.0429322 & 0.1721080 & 0.1170427 & 0.0907547 & 0.0691550 & 0.1252522 & 0.1090919\\
        FNO 
            & \textbf{0.0018643} & 0.0316177 & 0.1449448 & 0.0341974 & 0.0419211 & 0.0685995 & 0.0133442\\
        U-Net 
            & 0.0040814  
            & \textbf{0.0240380} 
            & \textbf{0.0251834} 
            & \textbf{0.0027415} 
            & \textbf{0.0076716} 
            & \textbf{0.0192060} 
            & \textbf{0.0030971}\\
        \bottomrule
        \end{tabular}
        \label{tab:auto-results-cylinder}
\end{table}


\clearpage

\begin{figure}
    \centering
    \includegraphics[width=15.5cm]{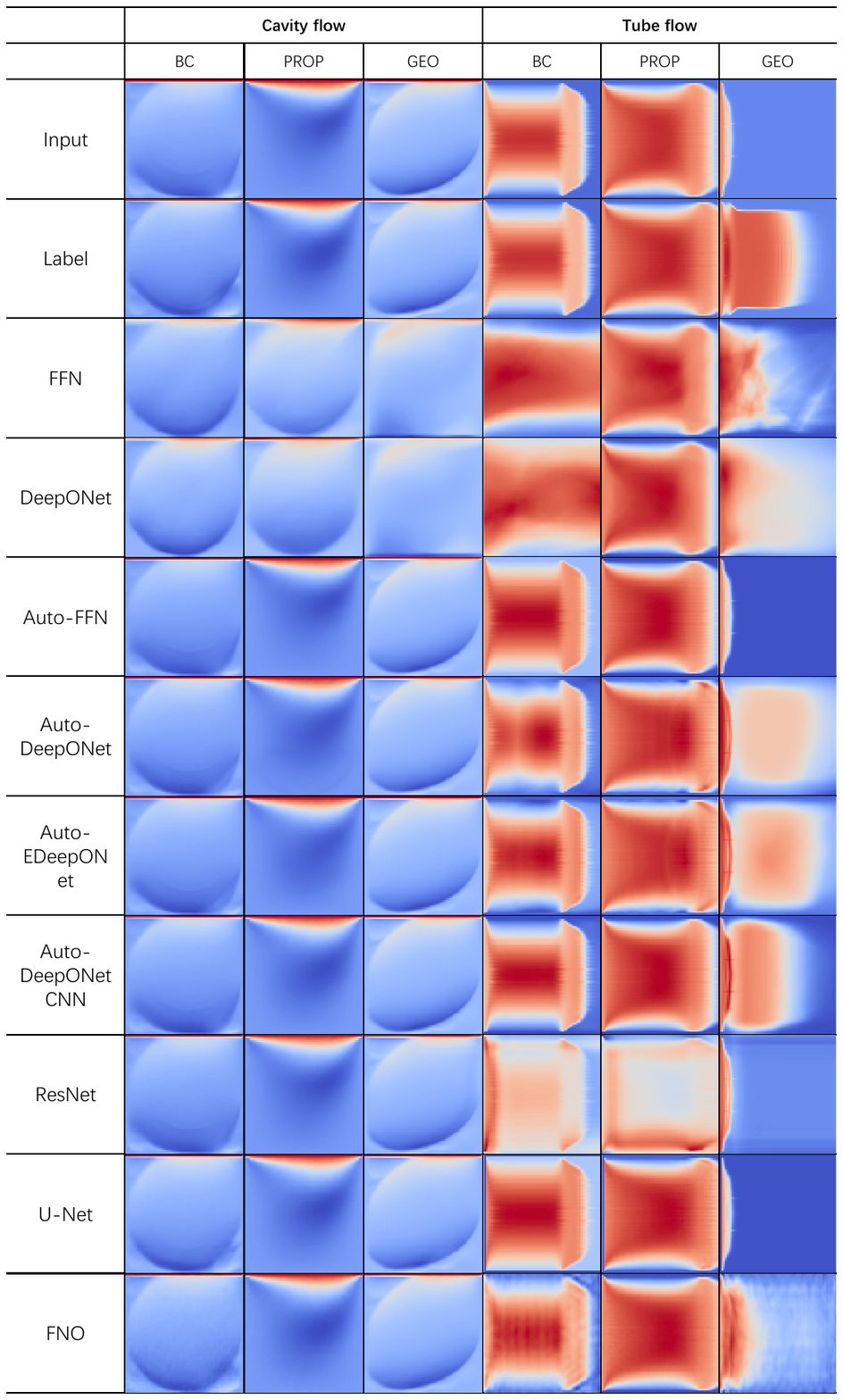}
    \caption{Prediction of the velocity field by the baseline models on the \textbf{cavity} and \textbf{tube} flow problems. The result of autoregressive models is the prediction of one forward propagation step. The input is not given to non-autoregressive models.}
    \label{fig:results-velocity-1}
\end{figure}

\begin{figure}
    \centering
    \includegraphics[width=15.5cm]{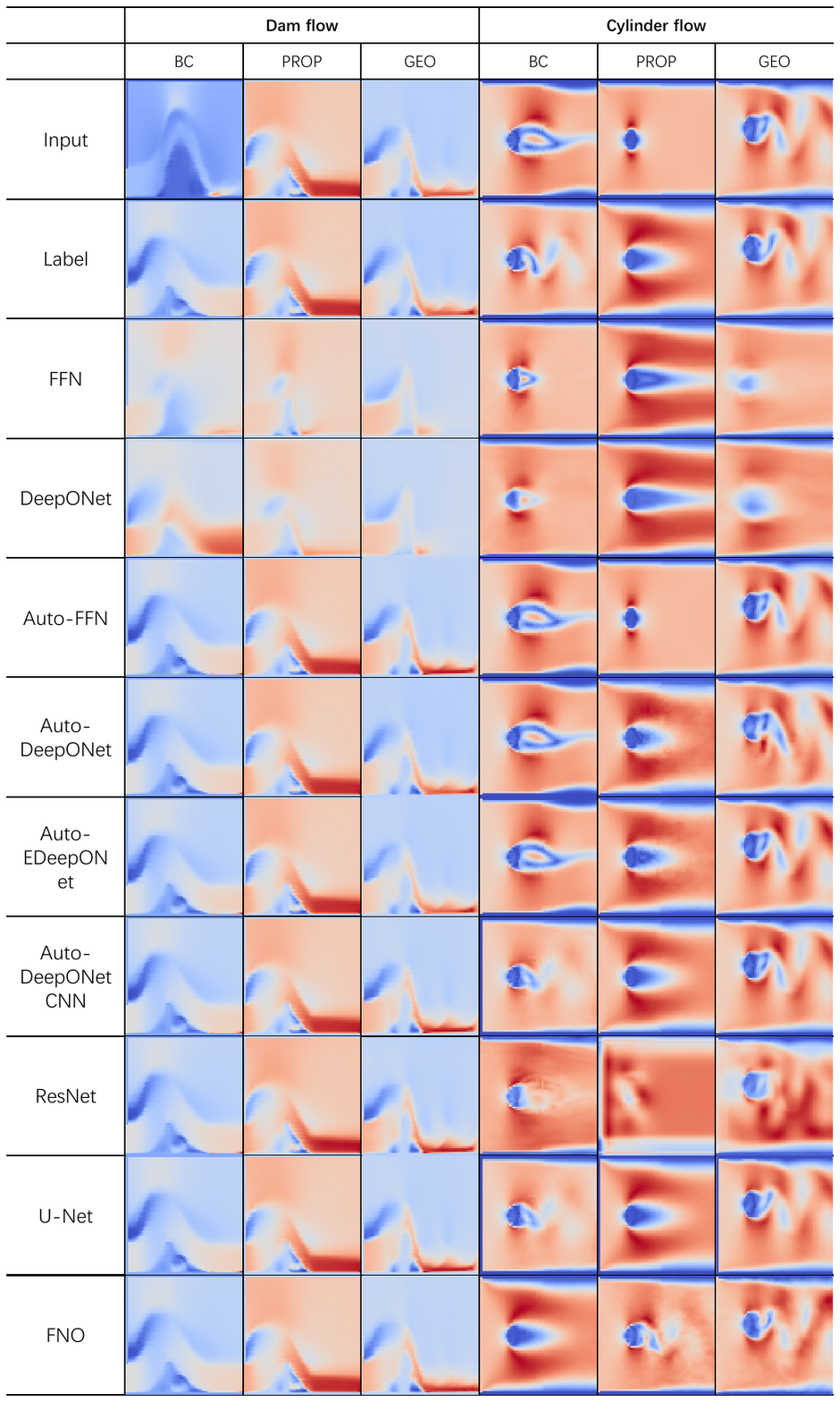}
    \caption{Prediction of the velocity field by the baseline models on the \textbf{dam} and \textbf{cylinder} flow problems. The result of autoregressive models is the prediction of one forward propagation step. The input is not given to non-autoregressive models.}
    \label{fig:results-velocity-2}
\end{figure}

\subsection{Qualitative Prediction Examples}

Here, we list some velocity field prediction results of all baseline models, shown in Figure \ref{fig:results-velocity-1} and Figure \ref{fig:results-velocity-2}.

\section{Hyperparameters of Baseline Neural Networks}
\label{sec:appendix-hyperparameter}

\subsection{DeepONet}

For the non-autoregressive DeepONet, we tried three activation functions, Tanh, ReLU, and GELU \citep{gelu} during hyperparameter search, and found that the validation NMSE and MSE of the model when using ReLU were significantly and consistently smaller than otherwise. This is different from some previous findings that indicate ReLU is worse at modeling flows \citep{pinn}. One reasonable explanation is that many existing works only consider periodic BCs and simpler flow problems, which is different from CFDBench. Moreover, our preliminary results show that in the cases where there is a circular flow, the model predicted a linear distribution instead, which indicates that the activation function does not capture the nonlinear characteristics well. To improve the activation function’s ability to model nonlinearity, we propose to normalize the input value of the activation function and add an activation function on the last layer of the branch net. We find that normalizing the input value of the activation function can significantly reduce NMSE, and removing activation functions after the last layer is better. 

\begin{figure}[H]
    \centering
    \includegraphics[width=0.5\linewidth]{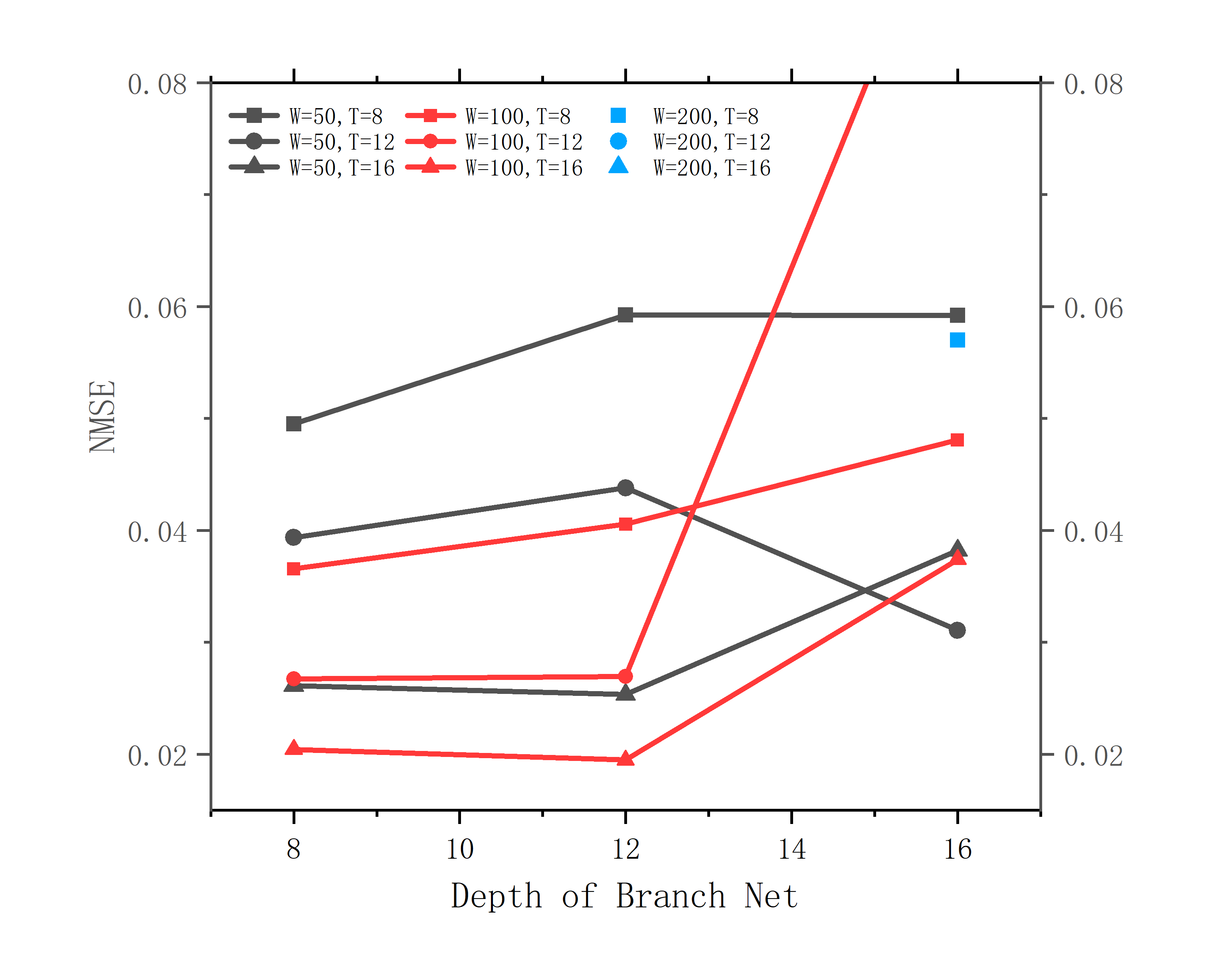}
    \caption{The validation loss of DeepONet using different hidden dimensions and numbers of FNO blocks on the cavity problem.}
    \label{fig:deeponet-hyperparameters}
\end{figure}

In this paper, unless stated otherwise, we use ReLU as the activation function and normalized the input value of the activation function, without the activation function for the last layer neurons. Additionally, preliminary trials show that DeepONet is unstable for inputs with large absolute values, so we normalized all the operating condition parameters.

The two sub-networks (branch net and trunk net) are feed-forward networks with constant width (output dimension). Therefore, the width and the depth (the number of fully connected layers) determine the number of parameters, and therefore the capacity of the model. We additionally conducted a hyperparameter search for the model’s width n (We follow the original DeepONet and set the width of the two sub-networks to the same value), the branch net’s depth $B$, and the trunk net’s depth $T$. The results are shown in Figure~\ref{fig:deeponet-hyperparameters}. The model displays severe overfitting when the parameters are too large. In the following sections, we used the parameters with the smallest validation NMSE, i.e., $n=100$, $B=12$, $T=16$. This results in 263,701 parameters.

\subsection{FNN and other models in the DeepONet Family}

For other models in the DeepONet family and the two FFNs, we generally wish to have their size be similar to other models. We start from the width and depth of the to sub-networks in the DeepONet model, and simply try different widths and depths aroudnd the same magnitude. The hyperparameters concerning the activation functions are the same as in DeepONet.

\subsection{U-Net}

In U-Net, each max-pooling layer is followed by two convolutional layers that double the number of channels. We consider searching the hidden dimension of the first convolutional layer ($d_2$) and the means of injecting operating condition parameters $\Omega$. We can either explicitly include $\Omega$ by adding it as addition channels of the input (as described in Section 3.2.3), or we can implicitly include it in the down-sampled hidden representations, i.e., we add a linear layer that projects $\Omega$ into the same shape as the encoder’s output and add it to that output. When the input explicitly contains $\Omega$, its dimensionality (i.e., number of channels) is $d_1=8$, the features are ($u\left(x,y\right)$, $v\left(x,y\right)$, $mask$, $u_\mathcal{B},\rho,\mu,h,w$) at every location on $(x,y)$, where $u$ and $v$ velocity along the $x$ and $y$ axis. When implicitly conditioning on $\Omega$, the input contains only the velocity field. which makes it $d_1=3$, which includes only $u, v$ and the mask. For $d_2$, it determines the dimensionality of each convolutional layer, thus, the intuition is that the larger $d_2$ is, the larger the model, and the stronger the learning ability. This is also the trend found by our hyperparameter search. However, a larger model also requires greater computational cost. On the other hand, we observe insignificant differences between the two ways to condition on the operating condition parameters. In the subsequent sections, we explicitly include $\Omega$ as additional input features (to make it more similar to the FNO structure that we use), and $d_2=12$. This results in 1,095,025 parameters.

\begin{figure}[H]
    \centering
    \includegraphics[width=0.5\linewidth]{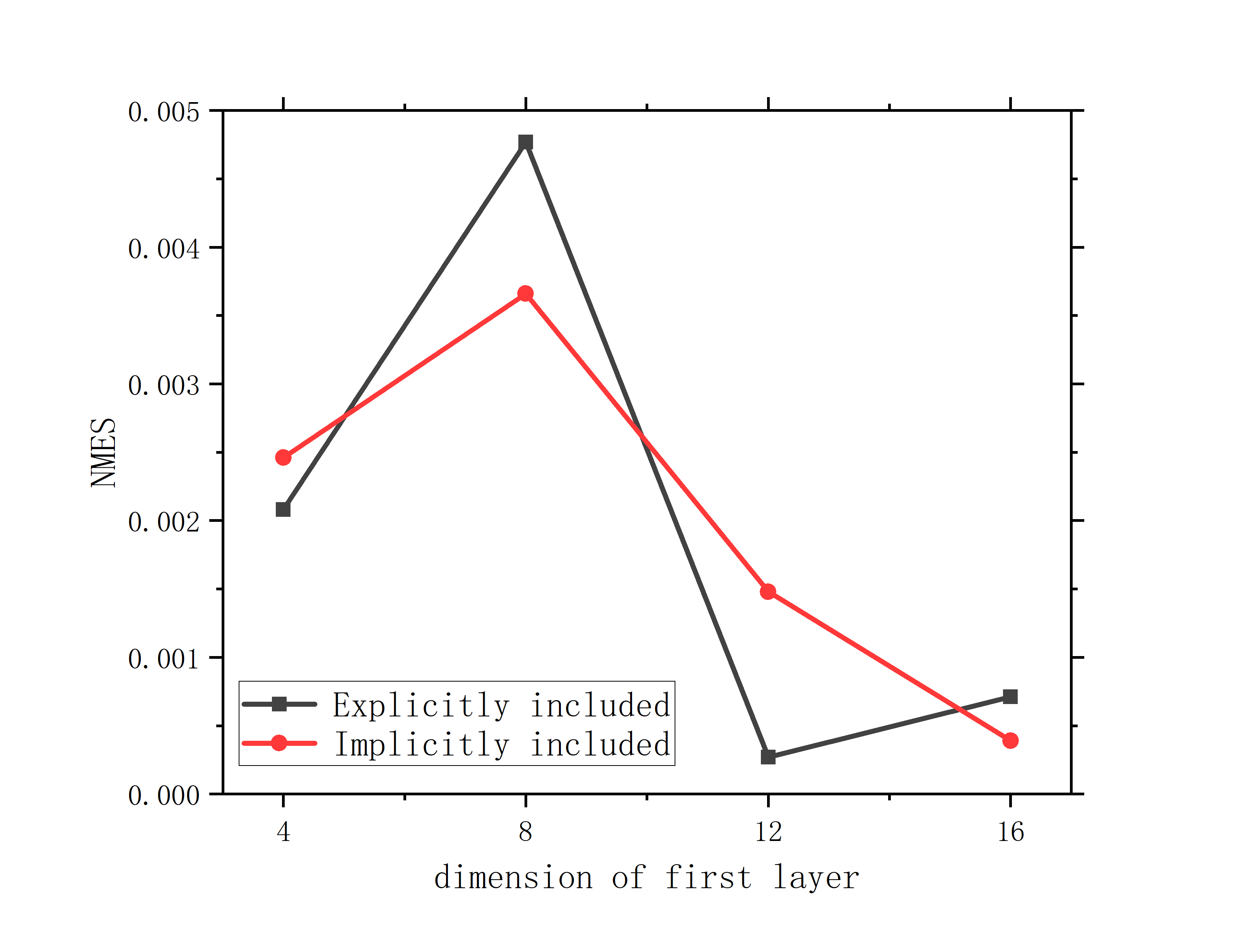}
    \caption{The validation loss of U-Net with different hidden dimensions and ways of conditioning on operating parameters on the cavity problem.}
    \label{fig:unet-hyperparameters}
\end{figure}

\subsection{FNO}

\begin{figure}[H]
    \centering
    \includegraphics[width=0.5\linewidth]{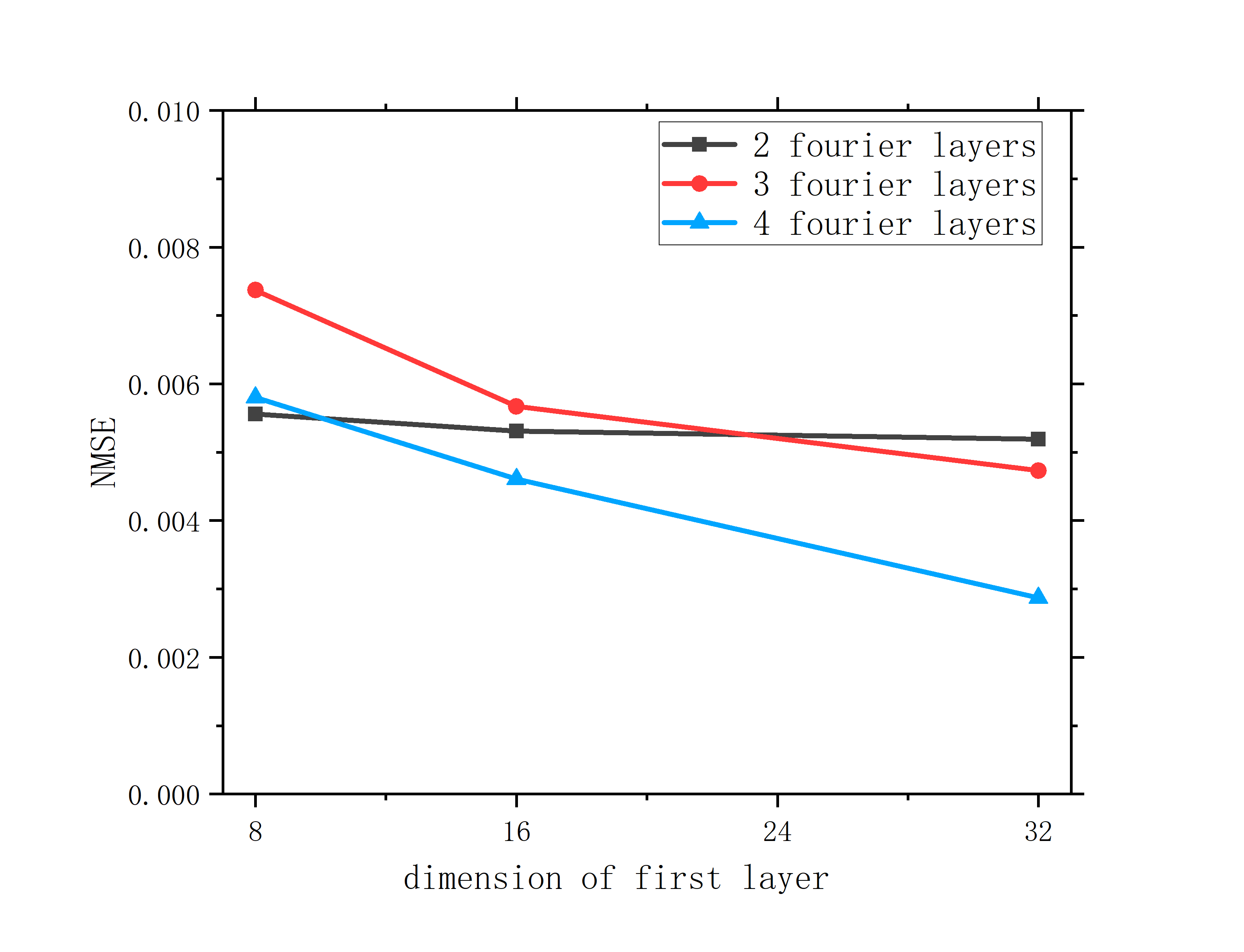}
    \caption{The validation loss of FNO using different hidden dimensions and number of FNO blocks on the cavity problem.}
    \label{fig:fno-hyperparameters}
\end{figure}

Since the structure of FNO is highly similar to ResNet, we search the same hyperparameters, namely, the number of FNO blocks ($d$) and the number of channels of the hidden representations ($h$). We choose to filter out high-frequency signals greater than 12 in the Fourier layer, which is used in the original paper. The results are shown in Figure~\ref{fig:fno-hyperparameters}. We observe that increasing both d and h can result in better validation loss, which is intuitive because both imply a greater number of parameters, which increases the capacity of the model. In order to ensure that the number of parameters of the baselines model is similar, such that the training and inference costs are similar, we choose to use $d=4$ and $h=32$, which results in 1,188,545 parameters.

\section{Data Processing}

\subsection{Interpolation Into Grids}

Before feeding the data to the neural networks for training, they require some pre-processing. Firstly, all data need to be discretized and interpolated into grids. For simplicity, we also keep the spatial domain to be $64\times 64$, so for different heights and widths, the size of the grid cells (denoted as $ \Delta x \times \Delta y$) are different. The value of each grid cell is set to be the average of the data points that fall into that cell, and if there is zero data points in a cell, its value is set to be the average of adjacent cells.\footnote{For multiple contiguous empty cells, we set their value iteratively from the boundaries of the empty region.}

Additionally, for BCs that are constants, we pad the tensor with one extra grid line. For instance, for the tube flow problem, we pad all grids on the top and bottom boundary with one line of zeros, resulting in a tensor with 66 rows and 64 columns.

\end{document}